\documentclass[sigconf]{acmart}
\renewcommand\footnotetextcopyrightpermission[1]{} % removes footnote with conference information in first column

\AtBeginDocument{%
  }

\setcopyright{acmlicensed}
\copyrightyear{2018}
\acmYear{2018}
\acmDOI{XXXXXXX.XXXXXXX}
\acmConference[Preprint]{-}{August, 2026}{}
\acmISBN{978-1-4503-XXXX-X/2018/06}

\usepackage{algorithm}
\usepackage{algorithmic}
\usepackage{multirow}
\usepackage[table]{xcolor}
\definecolor{darkblue}{RGB}{0,0,120}
\definecolor{darkgreen}{RGB}{0,100,0}

\begin{document}

\title{Fast Test-Time Refinement for Robust Learned Image Compression}


\author{Jiaming Liang}
\orcid{0000-0002-2510-9331}
\affiliation{%
  \institution{University of Macau}
  % \department{Faculty of Science and Technology}
  \city{Macau}
  \country{China}
}
\email{chinaliangjm@gmail.com}

\author{Chi-Man Pun}
\orcid{0000-0003-1788-3746}
\correspondingauthor
% \authornote{Corresponding author.}
\affiliation{%
  \institution{University of Macau}
  % \department{Faculty of Science and Technology}
  \city{Macau}
  \country{China}
}
\email{cmpun@um.edu.mo}

\author{Weisi Lin}
\orcid{0000-0001-9866-1947}
% \authornote{Corresponding author.}
\affiliation{%
  \institution{Nanyang Technological University}
  % \department{Faculty of Science and Technology}
  % \city{Macau}
  \country{Singapore}
}
\email{wslin@ntu.edu.sg}

% \correspondingauthor
% \affiliation{%

% \affiliation{%

% \affiliation{%

% \affiliation{%

% \affiliation{%

% \affiliation{%

% \affiliation{%

% \correspondingauthor
% \affiliation{%

\renewcommand{\shortauthors}{Jiaming Liang, Chi-Man Pun, Weisi Lin}

\begin{abstract}
Learned image compression (LIC) has demonstrated remarkable rate-distortion (RD) performance in benign settings. However, the high representational capacity endowed by deep neural networks (DNNs) comes at the expense of increased adversarial vulnerability. This hinders their adoption as trusted standardized codecs. Recent work has sketched test-time refinement (TTR) as a defense in gray-box scenarios, despite its original purpose of improving benign RD performance. Unfortunately, extensive iterations of TTR incur prohibitive overhead, while the robustness mechanism lacks theoretical understanding. Moreover, TTR has not been evaluated in white-box settings or against attacks beyond $\ell_{2}$-bounded rate and untargeted distortion objectives. To bridge these gaps, we present a systematic study. Our study reveals an \textbf{Asymmetric Adversarial Trajectory (AAT)} property in LIC systems: transitioning from adversarial to benign regions is significantly easier than the reverse process, where adversarial examples can often be roughly recovered within only 1–2 steps. We provide a two-dimensional \textbf{Tube Model} to explain this phenomenon. Based on AAT, we propose a \textbf{Fast Test-Time Refinement (FTTR)} framework for practical and robust LIC systems. We establish that the robustness arises from the contraction of adversarial regions induced by the \textbf{Input-as-Label} property of LIC systems, rather than from obfuscated gradients. Extensive evaluations with diverse strong adaptive attacks across multiple LIC systems demonstrate the promise of the proposed FTTR framework. The code is available at \url{https://github.com/chinaliangjiaming/FTTR.git}.
\end{abstract}



\begin{teaserfigure}
\centering
 \includegraphics[width=0.88\textwidth]{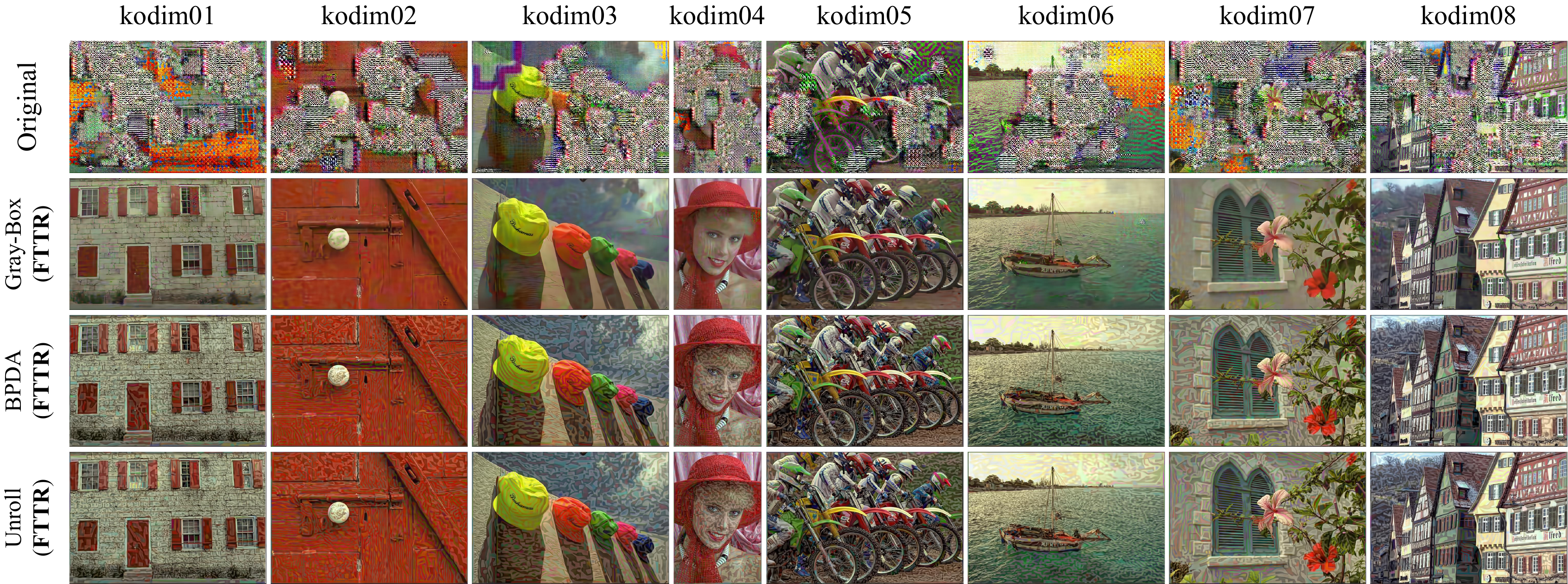}
  \caption{Reconstruction of examples generated by PGD (budget $16/255$, $400$ steps). \textit{Original}: attack and reconstruction on the same undefended LIC system $f$. \textit{Gray-Box}: attack on $f$ and reconstruction by FTTR-equipped $f$. \textit{BPDA} and \textit{Unroll}: white-box adaptive attacks on FTTR-equipped $f$, with STE approximation and exact gradients, respectively. Example based on DCAE~\cite{lu2025learned}.}
  \Description{caption}
  \label{fig:teaser}
\end{teaserfigure}


\maketitle

\section{Introduction}

\textbf{Background.} Learned image compression (LIC)~\cite{balle2017end, balle2018variational} incorporates deep neural networks (DNNs) to enhance representation capability, significantly reducing compression redundancy and achieving superior rate-distortion (RD) performance~\cite{zhang2026qarv++, jiang2026mlicv2}. However, the inherent high-dimensional vulnerability of DNNs~\cite{szegedy2013intriguing, goodfellow2015explaining} is also inherited by LIC systems, making them more susceptible to adversarial threats than conventional hand-crafted codecs such as JPEG~\cite{wallace1991jpeg}, JPEG 2000~\cite{skodras2002jpeg}, HEVC/H.265~\cite{sullivan2012overview}, and VVC/H.266~\cite{bross2021overview}.

Recent studies have demonstrated that imperceptible adversarial perturbations can induce rate collapse~\cite{kurihara2025efficient, wu2025adversarial}, reconstruction distortion~\cite{wu2025adversarial, kalmykov2026t}, or even manipulated reconstruction~\cite{chen2023toward, liang2026control}, as well as downstream degradation~\cite{sui2024transferable} in LIC systems. As a critical component of communication systems, compression requires strong reliability guarantees. In particular, standardized compression systems are expected to fully disclose all technical details, which exposes LIC systems to adversaries~\cite{chen2023toward}. Therefore, developing trustworthy LIC systems for secure deployment is imperative.

Despite this pressing need, research on robust LIC systems remains limited. One line of research improves the LIC robustness through adversarial training~\cite{chen2023toward, zhu2024attack, cao2024enhancing, wu2025adversarial, kurihara2025efficient}. However, in white-box settings, attackers can circumvent such defenses by generating adversarial examples directly against the adversarially trained model~\cite{tramer2020adaptive}. Another direction~\cite{song2024training, wu2025adversarial} keeps the model structure and parameters fixed, while mitigating adversarial inputs and optimizing latent representations at test time. Among them, static input transformations~\cite{song2024training} provide limited protection against attacks under white-box settings, as attackers can generate tailored perturbations by incorporating the transformation into the attack pipeline. In contrast, test-time refinement (TTR)~\cite{djelouah2019content}, as sketched in~\cite{wu2025adversarial}, provides dynamic protection against adversarial attacks through optimization, showing potential against white-box threats.

Specifically, TTR optimizes a purifier at test time via gradient descent based on the RD loss between the reconstructed image and the input image. The purifier directly or indirectly refines the latent and hyper-latent representations, and the optimized representations are then used for storage or transmission. TTR was originally proposed to improve benign RD performance at test time, particularly for domain adaptation scenarios, such as adapting codecs trained on natural images to screen content compression. \cite{wu2025adversarial} extends this idea by transferring the framework to adversarial inputs. Benefiting from its dynamic optimization property, TTR has the potential to make LIC systems robust against white-box attacks.

% TTR was originally proposed to iteratively refine the latent representations for improved benign RD performance.~\cite{wu2025adversarial} extends this idea by transferring the framework to adversarial inputs, demonstrating the promise of TTR for building robust LIC systems.
% This method fully exploits the Input-as-Label property of LIC systems and has been empirically shown to be promising.

\noindent\textbf{Gaps \& Challenges.} Unfortunately,~\cite{wu2025adversarial} only sketches TTR as a defense benchmark in the experiments without systematic investigation, leaving several important issues unaddressed. 
% However, test-time refinement has not been systematically investigated as an adversarial defense in~\cite{wu2025adversarial}, where it is only briefly presented as a benchmark without in-depth analysis, leaving several important issues unaddressed. 
\textbf{(I)} Conventional TTR requires a large number of optimization iterations, each involving backpropagation, making it computationally impractical. 
% The traditional TTR relies on extensive iterative optimization with per-iteration backpropagation, resulting in impractical computational costs.
\textbf{(II)} The theoretical foundation of TTR as a defense remains unexplored, leaving its robustness mechanism unclear.
% The theoretical justification is not established, and the robustness mechanism of TTR remains ambiguous. 
% ~\cite{wu2025adversarial} lacks theoretical justification, and the robustness mechanism of TTR remains ambiguous. 
\textbf{(III)} Existing evaluations of TTR as a defense have been confined to a narrow range of settings. \textbf{(III-a)} ~\cite{wu2025adversarial} considers only a gray-box setting, where TTR is disabled during attack generation and enabled only at test time. As a result, its robustness against white-box adaptive attacks remains unknown, casting doubt on whether the observed robustness stems from genuine effects or obfuscated gradients~\cite{athalye2018obfuscated}.
% Whether TTR's robustness genuinely arises from improved security or merely from obfuscated gradients remains unclear. Since~\cite{wu2025adversarial} disables adaptation during adversarial example generation but enables it during inference, TTR has not been validated against stronger adaptive attacks. 
% Whether the robustness of TTR stems from obfuscated gradients~\cite{athalye2018obfuscated} has not been verified.~\cite{wu2025adversarial} removes adaptation optimization during adversarial example generation while enabling it during inference. This mismatch raises concerns that the robustness may stem from obfuscated gradients.~\cite{wu2025adversarial} has not evaluated TTR against stronger adaptive attack paradigm, such as unrolled attacks~\cite{andrychowicz2016learning}, BPDA~\cite{athalye2018obfuscated}, and Monte Carlo sampling attacks~\cite{athalye2018obfuscated}. 
\textbf{(III-b)} TTR has primarily been evaluated under untargeted distortion and rate collapse attacks, with limited study on more recent threats such as semantic manipulation~\cite{liang2026control} and downstream task degradation~\cite{sui2024transferable}. 
% TTR has only been evaluated against untargeted distortion and rate collapse attacks, while its effectiveness against more recent threats, such as semantic manipulation~\cite{liang2026control} and downstream task degradation~\cite{sui2024transferable}, remains unexplored. 
\textbf{(III-c)} 
% $\ell_{2}$-bounded attacks typically concentrate perturbations on salient pixels, while $\ell_{\infty}$-bounded attacks distribute perturbations more uniformly across the image. 
TTR has only been evaluated under $\ell_{2}$ attacks, lacking a comprehensive assessment under other constraints.
% TTR has not been evaluated against attacks beyond the $\ell_{2}$ norm.
% Online Update only evaluates the robustness of test-time refinement against $\ell_{2}$-bounded perturbations. Its effectiveness against other commonly adopted attacks, particularly $\ell_{\infty}$-bounded attacks, remains unknown.

% \noindent\textbf{Insights.}

\begin{figure}[t]
    \centering
    \includegraphics[width=0.95\linewidth]{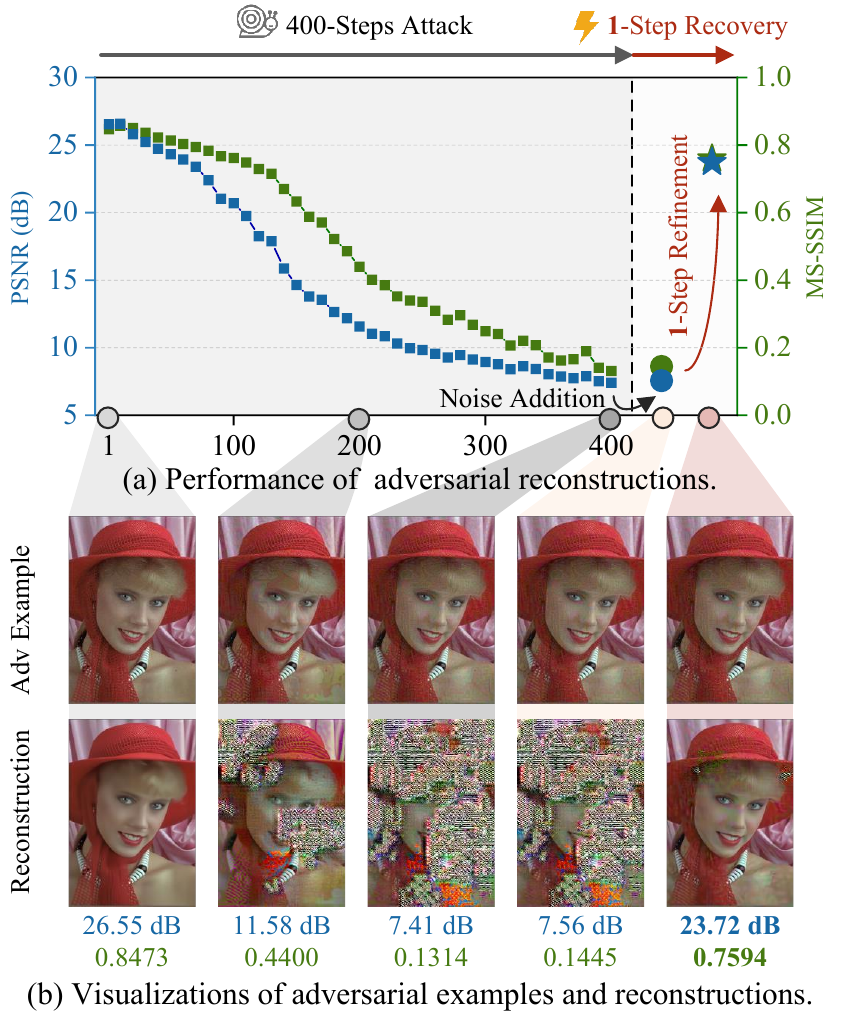}
    \caption{Our revealed Asymmetric Adversarial Trajectory (AAT) property of LIC systems: Generating effective adversarial examples against LIC systems requires numerous iterations, whereas their restoration to benign samples can typically be achieved within only 1–2 steps. The illustrated case is based on DCAE~\cite{lu2025learned} and PGD attack~\cite{madry2018towards}.}
    \label{fig:teaser}
\end{figure}

\noindent\textbf{Theory \& Methodology.} To bridge these gaps, we present a systematic study of TTR as an adversarial defense. 

\textbf{(I)} Our study uncovers an intriguing property of LIC systems, termed \textcolor{darkblue}{Asymmetric Adversarial Trajectory (AAT)} (Section~\ref{section: Asymmetric Adversarial Trajectory}). Specifically, we find that recovering adversarial examples of LIC systems often requires substantially fewer optimization steps than generating them. In most LIC systems, when the refinement strength is sufficient, the recovery process can be reduced to only 1–2 steps, as illustrated in Figure~\ref{fig:teaser}. We attribute the AAT property to the loss landscape shaped by the identity mapping property~\cite{liang2026control} of LIC systems, and further propose a two-dimensional \textcolor{darkblue}{Tube Model} (Section~\ref{section: Asymmetric Adversarial Trajectory}) to explain this phenomenon. Building upon the AAT property, we propose \textcolor{darkblue}{Fast Test-Time Refinement (FTTR)} (Section~\ref{section: Fast Test-Time Refinement}), an efficient framework of adversarial defense for LIC systems.

% One possible explanation of AAT is the flatness of the optimization landscape induced by the identity mapping property of LIC systems~\cite{liang2026control}.

\textbf{(II)} We theoretically show that TTR provides genuine robustness gains rather than obfuscated gradients. Unlike other deep learning systems, LIC systems possess the \textcolor{darkblue}{Input-as-Label} property (Section~\ref{section: White-Box Robustness Mechanism}), whereby the adversarial example itself acts as the target label during optimization. This unique property enables TTR to minimize the loss for each input, thereby contracting adversarial regions. Based on the AAT property, our FTTR is an efficient variant of TTR that achieves a favorable robustness-efficiency trade-off.
 % We theoretically show that FTTR contracts adversarial regions and transforms attack generation into a bilevel optimization problem whose solutions correspond to approximate saddle points of the induced objective landscape, substantially increasing attack complexity.

\textbf{(III)} We evaluate the robustness of FTTR under a wide range of settings (Section~\ref{section: Experiments and Results}). \textbf{(III-a)} Both white-box and gray-box settings are considered. In the white-box setting, we evaluate two strong adaptive attacker assumptions: Unroll~\cite{andrychowicz2016learning} and BPDA~\cite{athalye2018obfuscated}. \textbf{(III-b)} Our experiments cover diverse threats, including untargeted distortion, semantic manipulation, rate collapse, and downstream degradation. \textbf{(III-c)} Furthermore, we comprehensively evaluate the performance against both $\ell_2$- and $\ell_\infty$-bounded attacks. 

\noindent\textbf{Results.} Extensive experiments demonstrate that LIC systems equipped with our FTTR achieve strong white-box robustness with high efficiency. Specifically, under a 400-step $\ell_{\infty}$ white-box PGD attack with a budget of $16/255$, the average PSNR and bpp without FTTR are $9.90$ dB and $17.97$, respectively. In contrast, with \underline{1-step} FTTR, the average PSNR increases to $\mathbf{24.58}/\mathbf{25.00}$ dB ($+\mathbf{14.68}/\mathbf{15.10}$ dB) under BPDA/Unroll adversaries, while the average bpp decreases to $\mathbf{11.83}/\mathbf{11.57}$ ($-\mathbf{6.14}/\mathbf{6.40}$), respectively. Under a 1000-step $\ell_{2}$ white-box C\&W attack with $c=1000$, the average PSNR and bpp without FTTR are $7.59$ dB and $15.19$, respectively. In contrast, with \underline{1-step} FTTR, the average PSNR and bpp are $\mathbf{26.71}$ dB ($+\mathbf{19.12}$ dB) and $\mathbf{7.58}$ ($\mathbf{-7.61}$) under the BPDA adversary. Furthermore, under a white-box PGD-based LIC-triggered downstream classification attack with perturbation of $8/255$, our FTTR boosts the downstream classification accuracy from $0.4\%$ to $\mathbf{28.1\%}$ ($+\mathbf{27.7\%}$).

Our contributions can be summarized as follows:
\begin{itemize}
    \item To the best of our knowledge, this is the first systematic study of TTR as an adversarial defense.
    \item We reveal the AAT property of LIC systems, showing that recovering adversarial examples to benign regions requires few steps in most LIC systems. We propose the Tube Model to illuminate this intriguing phenomenon.
    \item Building upon the AAT property, we propose FTTR, an efficient framework for leveraging TTR in adversarial defense. 
    \item We theoretically show that the unique Input-as-Label property of LIC systems enables TTR to contract adversarial regions, thereby improving robustness. Our FTTR is an efficient variant of TTR that preserves strong robustness while substantially improving efficiency.
    \item Extensive experiments under diverse settings demonstrate that LIC systems equipped with our FTTR achieve strong robustness efficiently, highlighting its potential.
    % \item Our FTTR requires only 1–2 optimization steps, ensuring practical efficiency. We believe it can significantly advance the development of robust and practical LIC.
\end{itemize}

\section{Related Work}
\subsection{Learned Image Compression} LIC follows the transform coding framework, which consists of transform, quantization, and entropy coding. LIC integrates DNNs into the analysis–synthesis transform~\cite{balle2017end} to improve representation capacity and reduce redundancy, and into entropy modeling to accurately estimate latent distributions~\cite{balle2018variational}. Consequently, LIC achieves superior RD performance over conventional handcrafted compression codecs. Efforts to improve LIC models can be broadly grouped into three directions: optimization of the analysis–synthesis architecture~\cite{zou2022devil, feng2023nvtc, liu2023learned, li2024frequency, jiang2024llic, feng2025linear, zhang2026qarv++, chen2026adaptive}, quantization~\cite{alexandre2019learned, agustsson2020universally, zhou2020variable, guo2021soft, presta2025stanh}, and entropy modeling~\cite{minnen2020channel, he2021checkerboard, he2022elic, qian2022entroformer, li2024multirate, kim2024diversify, li2025learned, jiang2025mlic++, he2026practical}.

\subsection{Adversarial Threats on LIC}
Despite the impressive performance of LIC systems in benign settings, their reliance on DNNs makes them inherently susceptible to adversarial attacks. The adversarial threats to LIC systems can be divided as follows. (1) The first category causes rate collapse~\cite{liu2023manipulation, chen2023toward, yu2023backdoor, wu2025adversarial, kurihara2025efficient}, resulting in a significant increase in storage and transmission overhead. (2) The second category targets reconstruction fidelity. Depending on whether the adversarial reconstruction is manipulated toward a specific outcome, it can be further divided into untargeted reconstruction distortion~\cite{chen2023toward, yu2023backdoor, ma2024imperceptible, sui2024reconstruction, wu2025adversarial, kurihara2025efficient, kalmykov2026t} and semantic manipulation~\cite{chen2023toward, liang2026control}. (3) The last is downstream degradation~\cite{sui2024transferable}, where the adversary seeks to induce downstream task failures by manipulating reconstructions of LIC systems.

\subsection{Adversarial Defenses on LIC}
LIC systems are typically deployed in a white-box setting, where technical details are publicly disclosed and adversaries possess full knowledge of the system. Consequently, white-box robustness is of particular importance for LIC. Unfortunately, efforts to enhance LIC robustness remain scarce. Most existing work~\cite{chen2023toward, zhu2024attack, cao2024enhancing, wu2025adversarial, kurihara2025efficient} focuses on adversarial training~\cite{goodfellow2015explaining}. They alleviate vulnerability by incorporating adversarial examples during training. However, adversarial training struggles to eliminate adversarial vulnerability. Although it partially patches the vulnerabilities of the original LIC system, adversarial training may inadvertently introduce new vulnerabilities, causing the adversarial region to shift rather than contract. Attackers can then optimize against the adversarially trained model to craft adaptive examples that bypass the defense~\cite{tramer2020adaptive}.

In contrast, a limited number of studies focus on mitigating adversarial effects through input transformations, thereby projecting adversarial examples back onto the benign data manifold.~\cite{song2024training} places a static transformation before the LIC system and an inverse-transformation module after it. By comparing the reconstruction losses of multiple candidate transformation pairs, the method adopts the least-loss reconstruction as the final output. While effective in the black-box setting, the defense can be circumvented by adaptive attacks in the white-box setting, where all candidate transformation pairs are known to the adversary. Unlike this method, ~\cite{wu2025adversarial} pioneers the use of TTR based on the optimization for dynamic defense.

\subsection{Test-Time Refinement}
\label{section: Test-Time Refinement}
TTR~\cite{djelouah2019content} was originally proposed to improve in-distribution RD performance of LIC systems. Subsequently, TTR was extended to address the degradation of RD performance under distribution shifts, such as when LIC systems trained on natural images are applied to out-of-distribution data including illustrations or screen content images~\cite{tsubota2023universal, chen2025test}. With both the architecture and model parameters fixed, TTR performs iterative test-time optimization on the latent representation~\cite{djelouah2019content, guo2020variable, yang2020improving, lv2023dynamic, tsubota2023universal, chen2025test} and side information~\cite{guo2020variable, yang2020improving, chen2025test}, thereby obtaining a refined outcome.

However, TTR has long been confined to benign settings. ~\cite{wu2025adversarial} were the first to employ TTR as a defense. Unfortunately, TTR was only sketched as a gray-box defense benchmark in the experiments, leaving its potential for adversarial defense largely unexplored. Several critical questions remain unanswered, including the impracticality of multi-step optimization, the lack of a theoretical foundation for its robustness, and the absence of rigorous evaluations against diverse adaptive settings. In this work, we provide a systematic investigation of these issues and address them.

\begin{figure*}[t]
    \centering
    \includegraphics[width=0.90\linewidth]{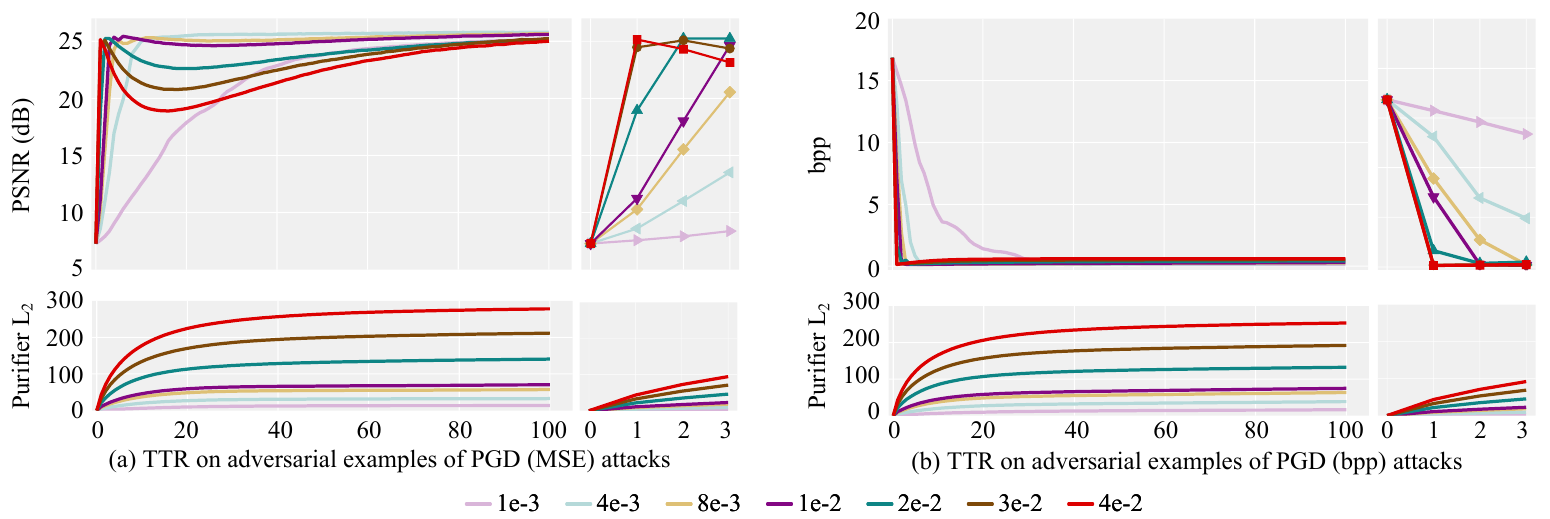}
    \caption{Performance of Adam-based TTR over iterations on PGD-generated MSE and bpp adversarial examples for DCAE. X-axis denotes the number of TTR iterations. The right subplot in each case provides a zoomed-in view of the first three TTR iterations. Colors indicate different Adam learning rates. \textit{Purifier $L_{2}$} denotes the $\ell_{2}$ norm of the corresponding purifier.}
    \label{fig:AAT}
\end{figure*}

\section{Theory and Methodology}
\subsection{Preliminaries}
\label{Preliminaries}
\textbf{LIC Systems.} For theoretical analysis, we abstract LIC systems into a general framework. An LIC system $f$ is composed of an encoder $E$, a quantizer $Q$, and a decoder $D$:
\begin{equation}
    f = D \circ Q \circ E,
\label{equation:definition_LIC}
\end{equation}
where $\circ$ indicates composition. The encoder $E$ encodes an input image $\boldsymbol{x}\sim\boldsymbol{\mathcal{X}}$ into a latent representation $\boldsymbol{y}$ and a hyper-latent representation $\boldsymbol{z}$: 
\begin{equation}
    \big[\boldsymbol{y}, \boldsymbol{z}\big] = E\big(\boldsymbol{x}\big),
\end{equation}
which are then quantized by decoder $Q$ into $\hat{\boldsymbol{y}}$ and $\hat{\boldsymbol{z}}$ for storage and transmission:
\begin{equation}
    ~\big[\hat{\boldsymbol{y}}, \hat{\boldsymbol{z}}\big] = Q\big(\boldsymbol{y}, \boldsymbol{z}\big).
\end{equation}
The decoder $D$ reconstructs the quantized latent representation $\hat{\boldsymbol{y}}$ into the reconstructed image $\hat{\boldsymbol{x}}$:
\begin{equation}
    \hat{\boldsymbol{x}} = D\big(\hat{\boldsymbol{y}}\big).
\end{equation}
The overall objective $\mathcal{L}$ consists of the bitrate $\mathcal{R}$ and distortion $\mathcal{D}$, which are balanced via a Lagrange multiplier $\lambda$:
\begin{equation}
\mathcal{L}=\underbrace{r\big(\hat{\boldsymbol{y}}\big) + r\big(\hat{\boldsymbol{z}}\big)}_{\mathcal{R}} + \lambda\cdot\underbrace{d\big(\boldsymbol{x}, \hat{\boldsymbol{x}}\big)}_{\mathcal{D}},
\label{equation:evaluation_loss}
\end{equation}
where $r$ measures the bit rate, and $d$ is a distortion measure.

\noindent\textbf{Test-Time Refinement.} To adapt to the test domain and further improve RD performance, TTR iteratively optimizes $\boldsymbol{y}$ and $\boldsymbol{z}$ via gradient descent on the RD objective at test time, yielding $\boldsymbol{y}^{*}$ and $\boldsymbol{z}^{*}$ as refined compression representations\footnote{Note that $r$ in Equation~\ref{equation:evaluation_loss} denotes the actual bit rate measured from the quantized representations during evaluation, whereas $r$ in Equation~\ref{equation:training_loss} represents the estimated bit rate used for optimization. }:
\begin{equation}
    \boldsymbol{y}^{*}, \boldsymbol{z}^{*} = {\arg\min}_{y,z}r\big(\boldsymbol{y}\big) + r\big(\boldsymbol{z}\big) + \lambda\cdot d\big(\boldsymbol{x}, \hat{\boldsymbol{x}}\big).
\label{equation:training_loss}
\end{equation}
$\boldsymbol{y}^{*}$ and $\boldsymbol{z}^{*}$ are then quantized by $Q$ for storage and transmission:
\begin{equation}
    \big[\hat{\boldsymbol{y}}^{*}, \hat{\boldsymbol{z}}^{*}\big] = Q\big(\boldsymbol{y}^{*}, \boldsymbol{z}^{*}\big),
\end{equation}
and decoded by $D$ to yield the refined reconstruction $\hat{\boldsymbol{x}}^{*}$:
\begin{equation}
    \hat{\boldsymbol{x}}^{*} = D\big(\hat{\boldsymbol{y}}^{*}\big).
\end{equation}
In this formulation, we do not explicitly define the purifier. Instead, it is implicitly embodied in the optimization process of Equation~\ref{equation:training_loss}. Specifically, the purifier can transform $\boldsymbol{x}$ into $\boldsymbol{x}^{*}$, thereby producing $\boldsymbol{y}^{*}$ and $\boldsymbol{z}^{*}$. Alternatively, it can directly optimize $\boldsymbol{y}$ and $\boldsymbol{z}$. Consequently, the purifier corresponds to $\boldsymbol{x}^{*}-\boldsymbol{x}$, $\boldsymbol{y}^{*}-\boldsymbol{y}$, and $\boldsymbol{z}^{*}-\boldsymbol{z}$.

\subsection{Threat Model}
\label{section:Threat Model}
The prior study~\cite{wu2025adversarial} presents TTR as a defense exclusively under a gray-box threat model, where TTR is enabled only at test time and excluded from the attack pipeline. As a result, adaptive attacks are not considered, leaving the robustness against fully informed adversaries unknown. This threat model is unrealistic and inconsistent with Kerckhoffs's principle~\cite{petitcolas2025kerckhoffs}, since the design of standardized compression codecs is publicly available. Therefore, besides the gray-box threat, we further consider the white-box threat.
% an adversary is assumed to have full knowledge of the defense and can therefore launch adaptive attacks. Therefore, we revisit the threat model and consider a more realistic white-box setting as follows.
% The prior study~\cite{wu2025adversarial} evaluates the robustness of TTR exclusively under a gray-box threat model. Specifically, TTR is not adaptively incorporated into the attack pipeline, but is enabled only at test time on the defended model. Consequently, the attacker is assumed to be unaware of the defense mechanism and unable to perform adaptive attacks, leading to a substantially incomplete assessment of TTR's actual robustness.
% In addition, this threat model is unrealistic and violates Kerckhoffs's principle~\cite{petitcolas2025kerckhoffs}. For standardized compression codecs, the technical details are expected to be fully disclosed. As a component of the codec, the defense mechanism should likewise be publicly available. Under such a setting, an adversary can launch adaptive attacks accordingly. Based on this, we adopt the following white-box threat model.

% \noindent\textbf{Adversary Assumptions.} 

Let $f_{d}$ denote the LIC system $f$ with an integrated defense. The adversary is assumed to have full knowledge of $f_{d}$ and can thus launch adaptive attacks $A\in\mathcal{A}$. For adaptive adversaries, we consider two attack assumptions, BPDA~\cite{athalye2018obfuscated} and Unroll~\cite{andrychowicz2016learning}. Detailed descriptions are provided in the appendix. The perturbations
\begin{equation}
    \boldsymbol{\delta}_{a}=A\big(f_{d}, \boldsymbol{x}\big)
\end{equation}
are constrained to the input, and the adversarial examples
 \begin{equation}
    \boldsymbol{x}_{\text{adv}}=\boldsymbol{x}+\boldsymbol{\delta}_{a}
\end{equation}
are processed by a trusted third party for compression, storage, and transmission. Man-in-the-middle (MITM) attacks are ruled out, preventing the adversary from tampering with the latent and hyper-latent representations. To ensure imperceptibility, perturbations $\boldsymbol{\delta}_{a}$ are constrained within an $\ell_{p}$ norm ball of $\epsilon_{a}$. In this paper, we consider the two widely studied cases, namely, $p=2$ and $p=\infty$. The adversarial objectives considered include compression rate $\mathcal{R}$, distortion $\mathcal{D}$, and downstream-task performance $\mathcal{P}$.

% The adversary is allowed to launch adaptive attacks $A\in\mathcal{A}$ only on the input sample $\boldsymbol{x}$, producing an imperceptible perturbation $\boldsymbol{\delta}^{a}=A(f^{d}, \boldsymbol{x})$ to generate the adversarial example $\boldsymbol{x}^{\text{adv}}$, which is used to interfere with the LIC system $f^{d}$. The compression and transmission processes are assumed to be handled by a trusted third party, and thus the adversary cannot perform man-in-the-middle (MITM) attacks.

\subsection{Asymmetric Adversarial Trajectory of LIC}
\label{section: Asymmetric Adversarial Trajectory}
\noindent\textbf{Test-Time Refinement for Adversarial Defense.}~\cite{wu2025adversarial} first introduced TTR as a defense benchmark for experiments. Specifically, they apply an iteratively refined purifier $\boldsymbol{\delta}_{d}^{*}$ to adversarial examples $\boldsymbol{x}_{\mathrm{adv}}$ for generating purified examples:
\begin{equation}
    \boldsymbol{x}_{\text{pur}}= \boldsymbol{x}_{\text{adv}} + \boldsymbol{\delta}_{d}^{*},
\end{equation}
where
\begin{equation}
    \boldsymbol{\delta}_{d}^{*} = {\arg\min}_{\boldsymbol{\delta}_{d}}\mathcal{L}_{\text{pur}},
\label{equation:optimal_purifier}
\end{equation}
and
\begin{equation}
    \mathcal{L}_{\text{pur}}=r\big(E\big(\boldsymbol{x}_{\text{adv}}+\boldsymbol{\delta}_{d}\big)\big) + \lambda\cdot d\big(\boldsymbol{x}_{\text{adv}}, f\big(\boldsymbol{x}_{\text{adv}}+\boldsymbol{\delta}_{d}\big)\big).
\label{equation:loss_pur}
\end{equation}
The purifier $\boldsymbol{\delta}_{d}^{*}$ is then propagated to indirectly optimize $\boldsymbol{y}$ and $\boldsymbol{z}$:
\begin{equation}
    \boldsymbol{y}^{*}, \boldsymbol{z}^{*}= E\big(\boldsymbol{x}_{\text{pur}}\big).
\end{equation}
It should be noted that latent and hyper-latent representations exhibit model-dependent scales, especially under adversarial settings, latent optimization may result in unstable updates. In contrast, input images are normalized, and input optimization provides stable parameterization. Therefore, we follow this strategy in this work.

\noindent\textbf{Asymmetric Adversarial Trajectory.} We begin our discussion with the gray-box threat model. To solve Equation~\ref{equation:optimal_purifier},~\cite{wu2025adversarial} performs gradient descent via Adam~\cite{kingma2015adam} with a small learning rate $\alpha=0.01$ over multiple iterations. While numerous iterations are justified when optimizing benign inputs, they are unnecessary for the intent of defense. To illustrate this, we generate the MSE and bpp adversarial examples from \texttt{kodim04.png} in the Kodak~\cite{kodak1993} by $400$ iterations of PGD with a budget of $16/255$ and a step size of $2/255$. TTR is then performed for $100$ steps with Adam under different $\alpha$. DCAE~\cite{lu2025learned} is adopted as the tester. Results are shown in Figure~\ref{fig:AAT}.

\begin{figure}
    \centering
    \includegraphics[width=0.88\linewidth]{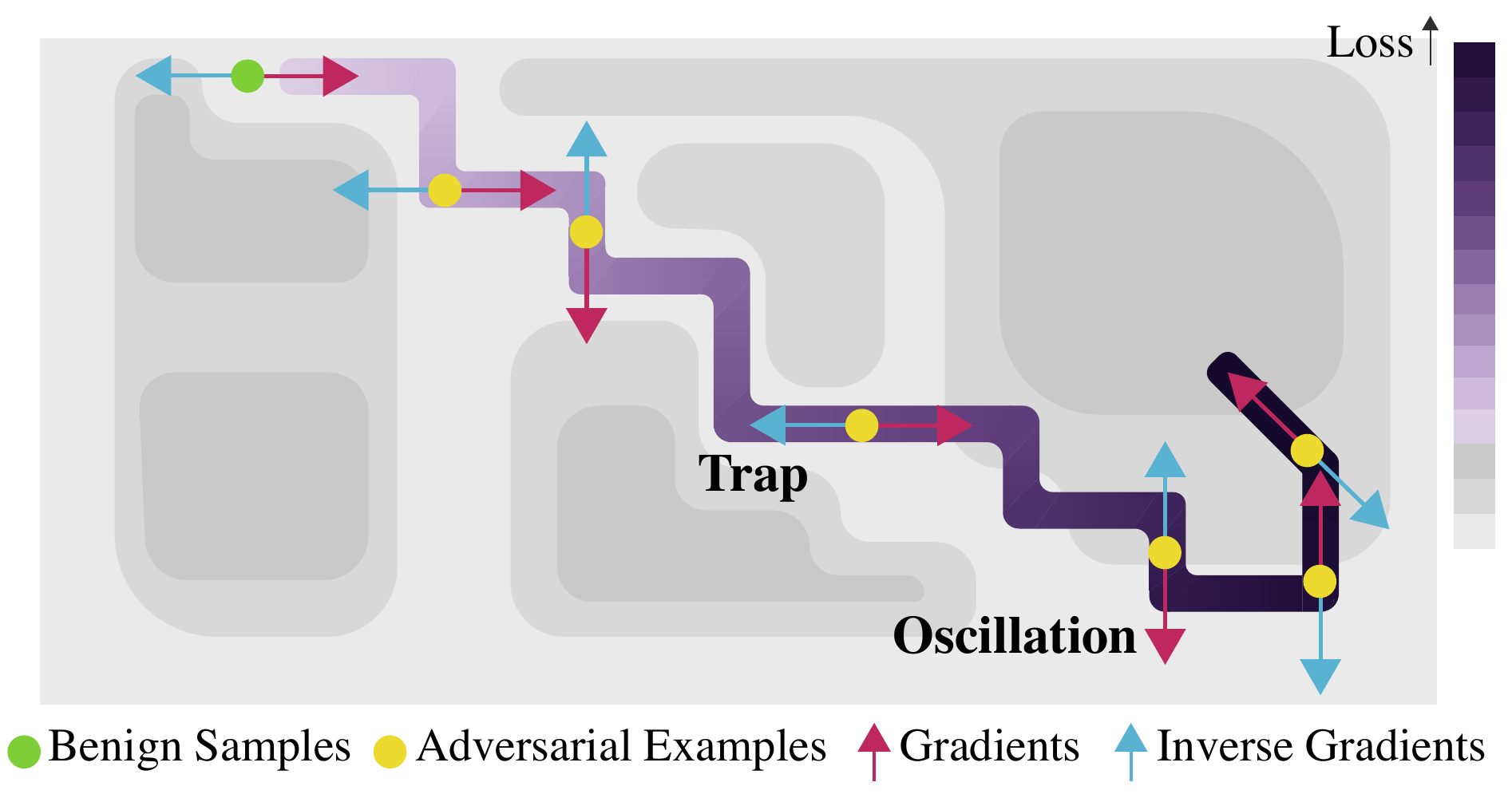}
    \caption{Hypothesized Tube Model for the two-dimensional structure of adversarial regions in LIC systems.}
    \label{fig:tube_model} 
\end{figure}

From the figure, TTR with Adam under different learning rates $\alpha$ converges to similar performance within 100 steps, indicating that learning rates $\alpha$ have limited impact given sufficient iterations. However, under few-step optimization, learning rates $\alpha$ play a pivotal role for defense. Small $\alpha$ yields stable but slow convergence, whereas large $\alpha$ reaches near-convergent performance within $1$–$2$ steps, albeit with transient instability followed by recovery. Although this observation is based on $\boldsymbol{x}_{\text{adv}}^{(400)}$, extensive experiments find it consistently holds for adversarial examples generated at arbitrary time steps $t$. Therefore, while generating adversarial examples in LIC systems is strenuous, restoring them to benign samples is remarkably effortless. We refer to this intriguing property of LIC systems as \textcolor{darkblue}{Asymmetric Adversarial Trajectory (AAT)}. 

This naturally leads to the following question. (Q1) Why can large $\alpha$ reach near-convergent performance in even a single step? (or Why does AAT exist?) (Q2) Why does the performance subsequently drop sharply before recovering? The answer to (Q2) is relatively straightforward. Adam is designed for iterative convergence and inherently assumes a large number of iterations. Although its first update aligns with the gradient direction (see Equation~\ref{equation:reduction_sign_gradient_descent}), subsequent updates in the early iterations may deviate from the gradient. As a result, with a large $\alpha$ and low iterations, the optimizer may step in an incorrect direction, leading to degraded performance. Regarding (Q1), we next provide a detailed analysis.

\noindent\textbf{Hypothesized Tube Model.} When $\boldsymbol{\delta}_{d}^{(0)}$ is initialized as $\boldsymbol{0}$ and the stability constant $\epsilon\ll |\nabla_{\boldsymbol{\delta}_{d}}\mathcal{L}_{\text{pur}}|$, the first Adam update reduces to a sign gradient descent step with $\alpha$ (see Appendix for the deduction):
\begin{equation}
    \boldsymbol{\delta}^{(1)}_{d}=-\alpha\cdot\text{sgn}\big(\nabla_{\boldsymbol{\delta}_{d}^{(0)}}\mathcal{L}_{\text{pur}}\big).
\label{equation:reduction_sign_gradient_descent}
\end{equation}
In other words, at the first TTR iteration, Adam moves an $\ell_{\infty}$ distance of $\alpha$ along the negated gradient direction, i.e., away from the adversarial region. Combined with the revealed AAT property of LIC systems, this suggests that for adversarial examples $\boldsymbol{x}_{\mathrm{adv}}^{(t)}$ generated at arbitrary attack steps $t$, they can escape the adversarial region and return to the benign region via a 1-step sign gradient descent with a sufficiently large $\alpha$. This indicates that the adversarial region has extremely limited thickness along the negated gradient direction. Otherwise, a single-step update would be insufficient to escape the region if its thickness were large.

From a two-dimensional perspective, this analysis suggests that the adversarial region of LIC systems is likely to exhibit a thin, highly curved, and tube-like geometric structure. We refer to this hypothesized two-dimensional geometry of the adversarial region of LIC systems as \textcolor{darkblue}{Tube Model}. As illustrated in Figure~\ref{fig:tube_model}, due to the highly curved geometry of the adversarial region, linear updates of the adversarial perturbation $\boldsymbol{\delta}_a$ fail to reach effective adversarial examples within a few iterations, and instead require numerous small-step updates that progressively follow the loss-increasing trajectory. In contrast, for most points, one or two large-step updates along the negated gradient direction are sufficient to escape the adversarial region, which is consistent with the AAT property revealed above. Moreover, the Tube Model also accounts for the previously observed adversarial behaviors in LIC systems. For example, PGD$^{2}$-GSM~\cite{liang2026control} exhibits pronounced oscillations when attacking with large step sizes. This can be attributed to step sizes exceeding the local straight-segment length of the tube-like adversarial region, leading to overshooting into benign regions and consequently inducing oscillatory dynamics. In summary, Tube Model provides a new geometric perspective on the adversarial region of LIC systems.
% According to AAT, a single large sign gradient descent step is sufficient to escape the adversarial region. Since this observation holds for $\boldsymbol{x}_{\mathrm{adv}}^{(t)}$ at arbitrary time steps $t$, it suggests that adversarial examples consistently resides near the boundary of the adversarial region. Therefore, the adversarial region in LIC systems is likely highly anisotropic, exhibiting a very small thickness along the gradient-descent direction. From a two-dimensional perspective, it can be viewed as a long and winding tube, as in Figure~\ref{fig:tube_model}. We term this hypothesized two-dimensional geometric structure of adversarial region the \textcolor{darkblue}{Tube Model}.

\begin{figure}[t]
    \centering
    \includegraphics[width=0.98\linewidth]{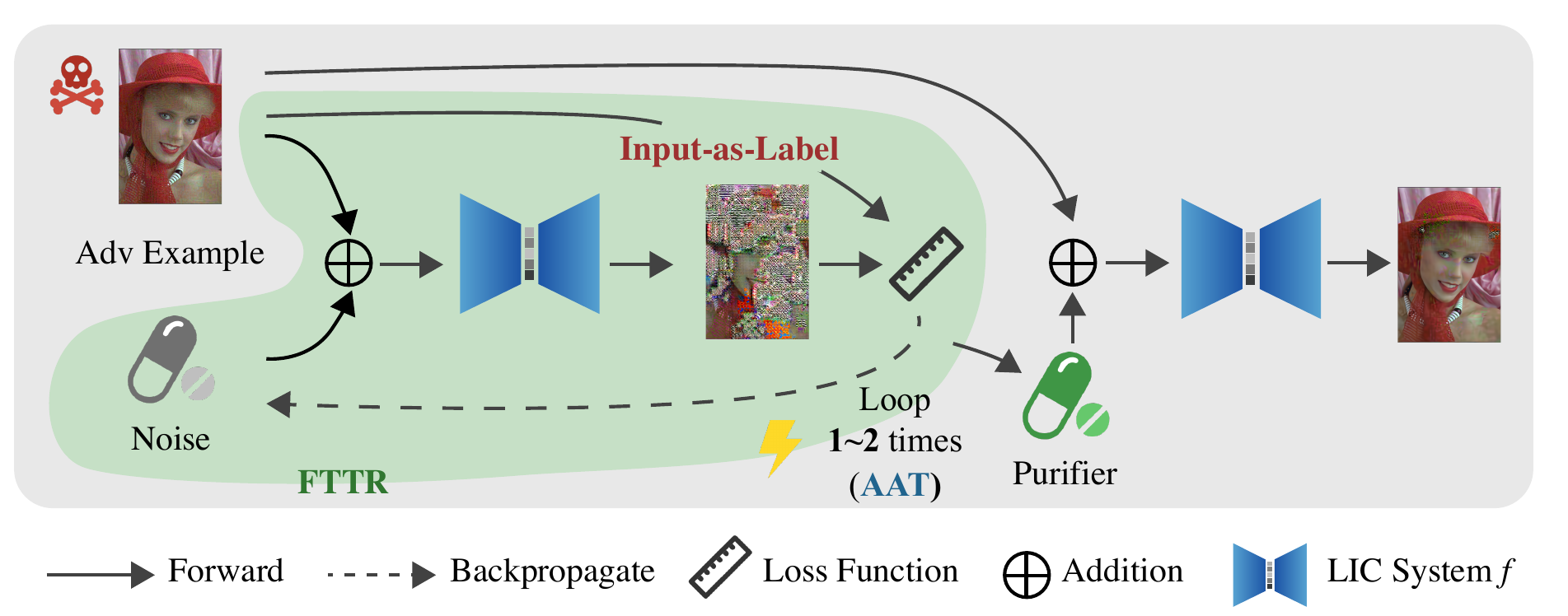}
    \caption{Pipeline of our FTTR for adversarial defense.}
    \label{fig:pipeline_FTTR}
\end{figure}

% The conventional TTR optimizes representations within the benign region, which is typically flat, especially for LIC systems behaving as near-identity mappings. This requires extensive optimization steps. In contrast, adversarial defense aims to move adversarial examples from the adversarial region to the benign region, which is fundamentally different from the benign setting. Excessive optimization steps lead to unnecessary computational overhead and impracticality.

\subsection{Fast Test-Time Refinement for Defense}
\label{section: Fast Test-Time Refinement}
Therefore, numerous iterations are unnecessary when TTR is deployed for defense purposes. In practice, TTR for defense typically requires only a few iterations, or even a single-step update. Motivated by this insight, we redesign TTR into a lightweight defense framework, termed \textcolor{darkblue}{Fast Test-Time Refinement (FTTR)}.

\begin{algorithm}[tb]
\caption{Compression with Fast Test-Time Refinement (FTTR)}
\label{alg:FTTR}
\textbf{Input}: Adversarial example $\boldsymbol{x}_{\text{adv}}$, LIC system $f$\\
\textbf{Parameter}: Radius $s$ of purifier $\boldsymbol{\delta}_{d}$, iteration number $K$\\
\textbf{Output}: Optimized $\boldsymbol{y}^{*}$ and $\boldsymbol{z}^{*}$
\begin{algorithmic}[1] %[1] enables line numbers
\STATE Initialize $\boldsymbol{\delta}_{d}^{(0)}\sim \text{Uniform}(-s,s)$
\FOR{$k=1$ to $K$}
\STATE $\boldsymbol{g}^{(k)}=\text{sgn}\big(\nabla_{\boldsymbol{\delta}_{d}^{(k-1)}}\mathcal{L}_{\text{FTTR}}^{(k)}\big)$
\STATE $\boldsymbol{\delta}_{d}^{(k)}=\text{Clip}_{(-s,s)}\big(\boldsymbol{\delta}_{d}^{(k-1)}-s\cdot K^{-\text{sgn}(k-1)}\cdot\boldsymbol{g}^{(k)}\big)$

\ENDFOR
\STATE\textcolor{darkgreen}{\# The following conditional branch is only used in practice to \# ensure that benign performance is not degraded.}
\IF{$\mathcal{L}_{\text{FTTR}}(\boldsymbol{x}_{\text{adv}},\boldsymbol{\delta}_{d}^{(K)})<\mathcal{L}_{\text{FTTR}}(\boldsymbol{x}_{\text{adv}},\boldsymbol{0})$}
\STATE \textbf{return} $E\big(\boldsymbol{x}_{\text{adv}}+\boldsymbol{\delta}_{d}^{(K)}\big)$
\ELSE
\STATE \textbf{return} $E\big(\boldsymbol{x}_{\text{adv}}\big)$
\ENDIF

\end{algorithmic}
\end{algorithm}

FTTR initializes the purifier $\boldsymbol{\delta}_{d}^{(0)}$ with uniform noise rather than zeros. This initialization incurs negligible overhead while nudging adversarial examples away from the adversarial region. The purifier is then refined through one or a few optimization steps. The refinement loss follows Equation~\ref{equation:loss_pur} to jointly optimize multiple objectives. Specifically, to defend against compression rate collapse:
\begin{equation}
    \mathcal{L}_{\text{BPP}}^{(k)}=r\big(E\big(\boldsymbol{x}_{\text{adv}}+\boldsymbol{\delta}_{d}^{(k-1)}\big)\big),
\end{equation}
For reconstruction quality, both the MSE loss:
\begin{equation}
    \mathcal{L}_{\text{MSE}}^{(k)}=\text{MSE}\big(f\big(\boldsymbol{x}_{\text{adv}}+\boldsymbol{\delta}_{d}^{(k-1)}\big),\boldsymbol{x}_{\text{adv}}\big),
\end{equation}
and the MS-SSIM loss:
\begin{equation}
    \mathcal{L}_{\text{MS-SSIM}}^{(k)}=\text{MS-SSIM}\big(f\big(\boldsymbol{x}_{\text{adv}}+\boldsymbol{\delta}_{d}^{(k-1)}\big),\boldsymbol{x}_{\text{adv}}\big)
\end{equation}
are adopted. When the refinement additionally aims to defend downstream attacks, an extra term:
\begin{equation}
    \mathcal{L}_{\text{opt}}^{(k)}=J\big(\boldsymbol{x}_{\text{adv}}, \boldsymbol{\delta}_{d}^{(k-1)}\big)
\end{equation}
is incorporated, where $J$ is the criterion of the downstream task. Accordingly, the $k^{\text{th}}$-step iterative objective is summarized as:
\begin{equation}
    \mathcal{L}_{\text{FTTR}}^{(k)}=\mathcal{L}_{\text{BPP}}^{(k)}+\lambda_{1}\cdot\mathcal{L}_{\text{MSE}}^{(k)}-\lambda_{2}\cdot\mathcal{L}_{\text{MS-SSIM}}^{(k)} + \lambda_{3}\cdot\mathcal{L}_{\text{opt}}^{(k)}.
\end{equation}
To prevent excessive distortion, the purifier is constrained within an $\ell_{\infty}$ ball of radius $s$. The first optimization step performs gradient descent along the negated sign gradient direction with step size $s$. Owing to the AAT property of LIC systems, this single large-step update is sufficient in most cases. When iteration $K>1$, subsequent steps continue with a reduced step size of $s/K$. The complete procedure is summarized in Algorithm~\ref{alg:FTTR}, while the overall pipeline is illustrated in Figure~\ref{fig:pipeline_FTTR}.

\subsection{White-Box Robustness Mechanism}
\label{section: White-Box Robustness Mechanism}
The preceding analysis focuses on the gray-box setting, where refinement is disabled during attacks but enabled at test time. As discussed in Section~\ref{section:Threat Model}, white-box robustness is crucial for LIC systems. Therefore, we further examine the white-box robustness.

For most deep learning tasks, achieving white-box robustness remains challenging, because the groundtruth of adversarial examples are inaccessible. For instance, in classification systems, an image of \texttt{dog} may be perturbed to be misclassified as \texttt{kangaroo}. The defender has no access to its true label \texttt{dog}, making recovery to a benign state intractable. Consequently, existing defenses rely on transformations, adversarial training, or related techniques to alleviate adversarial effects. Unfortunately, while they mitigate the original adversarial vulnerabilities, they often introduce new ones. In other words, they merely induce a shift of the adversarial region, rather than shrinking or eliminating it, which lies at the heart of the difficulty in achieving white-box robustness.

\begin{figure}[t]
    \centering
    \includegraphics[width=0.98\linewidth]{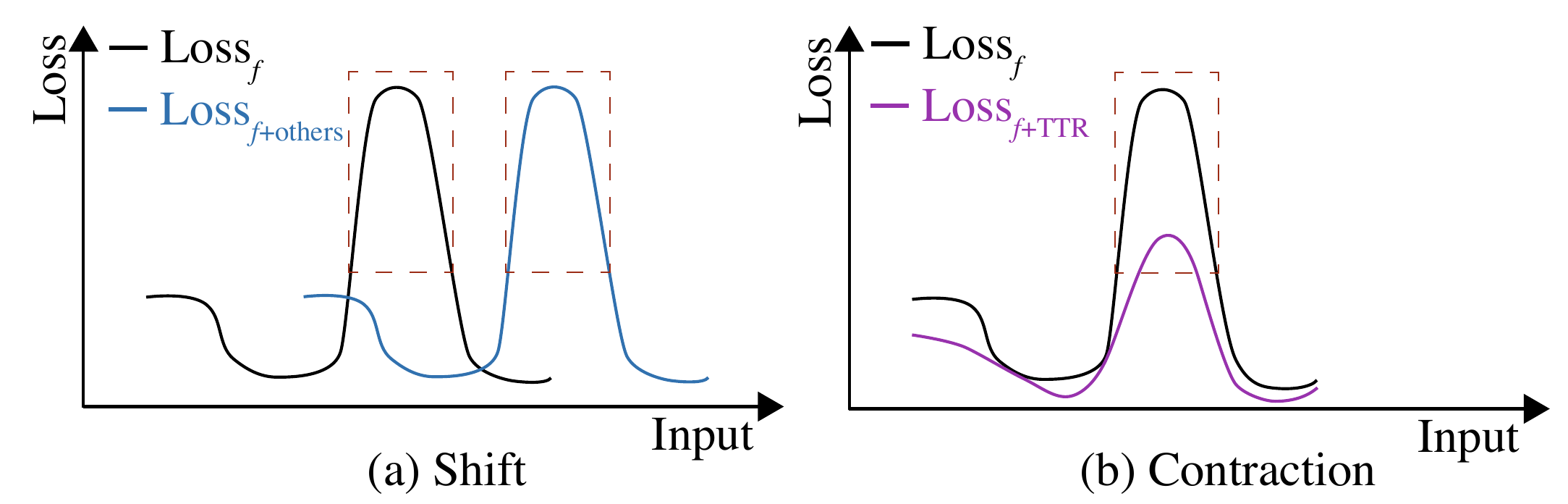}
    \caption{Illustrations on the shift and contraction of adversarial regions. TTR leverages the Input-as-Label to contract the adversarial region, enhancing white-box robustness.}
    \label{fig:shift_contraction_adversarial_region}
\end{figure}

\begin{figure*}[t]
    \centering
    \includegraphics[width=0.95\linewidth]{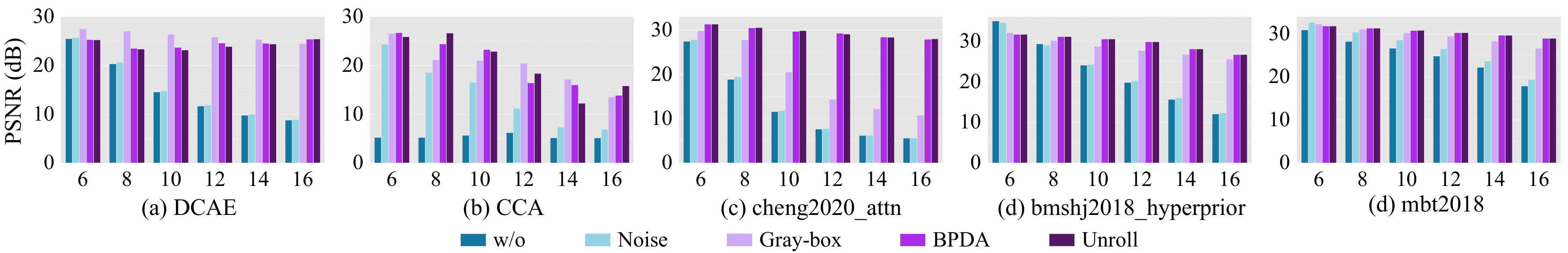}
    \caption{PSNR (dB) of PGD-attacked reconstructions under MSE. X-axis indicates perturbation budgets. \textit{w/o}: attack/inference on undefended $f$. \textit{Noise}: attack on undefended $f$, inference with noisy input. \textit{Gray-box}: attack on undefended $f$, inference with FTTR. \textit{BPDA}/\textit{Unroll}: attack/inference with FTTR (see Appendix for details). The same notation applies hereafter.}
    \label{fig:experiment_PGD_MSE_PSNR}
\end{figure*}

\begin{figure*}[t]
    \centering
    \includegraphics[width=0.95\linewidth]{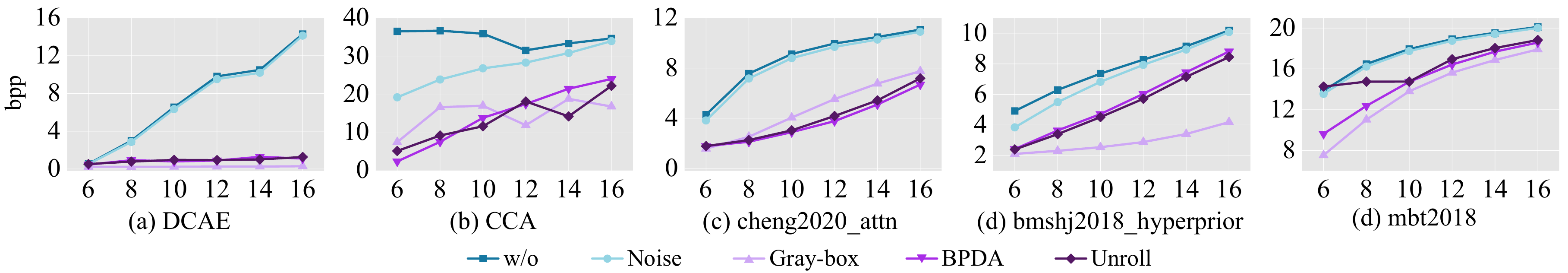}
    \caption{bpp of PGD-attacked reconstructions under bpp.}
    \label{fig:experiment_PGD_bpp_bpp}
\end{figure*}

\begin{figure}
    \centering
    \includegraphics[width=0.95\linewidth]{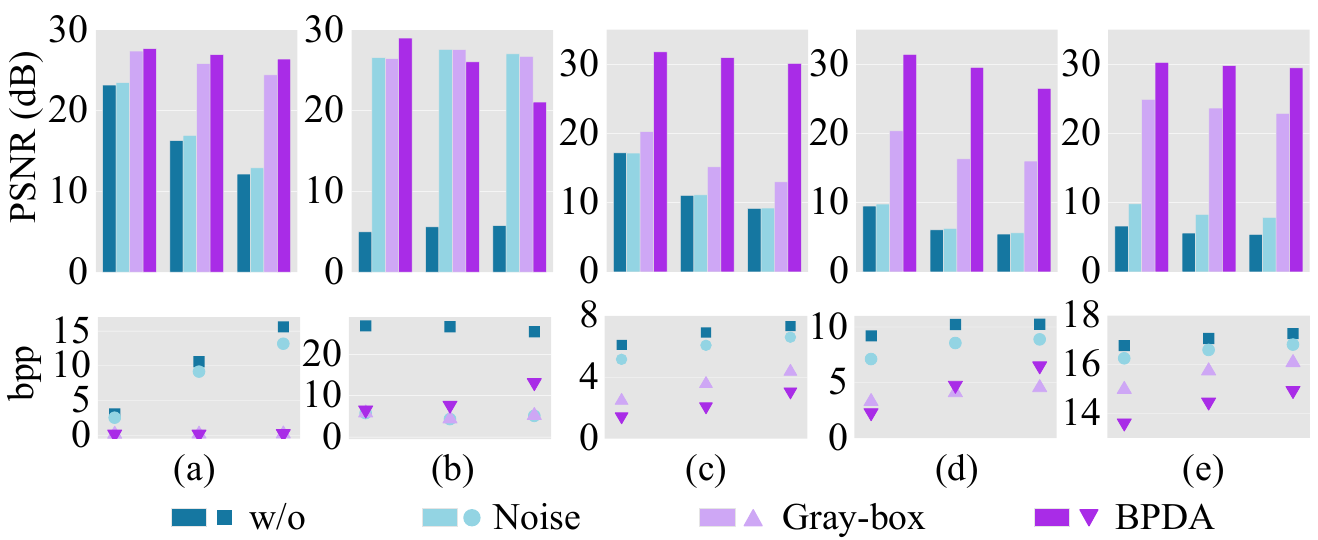}
    \caption{PSNR and bpp of C\&W-attacked reconstructions under joint MSE and bpp. Order of (a)-(e) follows Figure~\ref{fig:experiment_PGD_MSE_PSNR}.}
    \label{fig:CW_PSNR_BPP}
\end{figure}

\begin{figure*}
    \centering
    \includegraphics[width=0.95\linewidth]{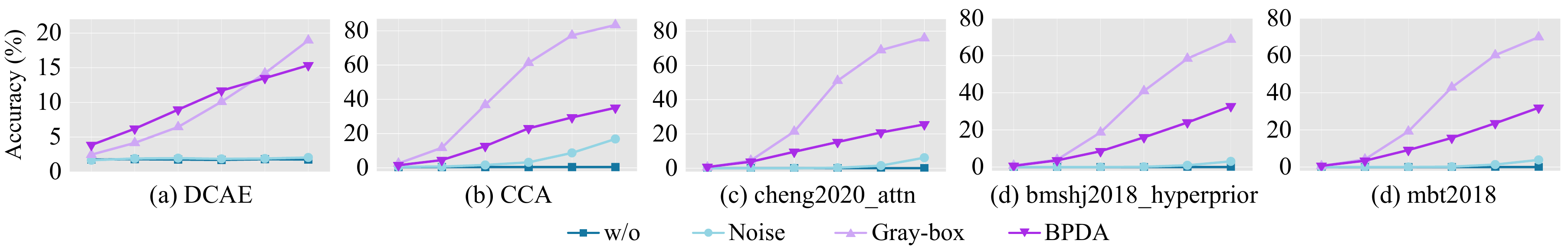}
    \caption{Classification accuracy (\%) of LIC-triggered downstream attacks.}
    \label{fig:classification}
\end{figure*}

% \begin{figure*}
%     \centering
%     \includegraphics[width=0.95\linewidth]{Figures/benign_vesion2.pdf}
%     \caption{PSNR (dB) and bpp of FTTR on benign samples.}
%     \label{fig:benign}
% \end{figure*}

In contrast, LIC systems differ fundamentally from them. Under the control of a trusted third party over compression, storage, and transmission, the ground truth for any adversarial example is inherently available, since the reconstruction target is the adversarial example itself. We refer to this as the \textcolor{darkblue}{Input-as-Label} property of LIC systems. TTR can serve as a white-box defense precisely because it exploits the Input-as-Label property of LIC systems. With an unlimited number of iterations, it is easy to see that TTR ensures that the loss can be reduced for any adversarial example, or at least remains non-increasing (because the trivial solution can be $\boldsymbol{\delta}_{d}^{*}=0$):
\begin{equation}
    \forall\boldsymbol{x}_{\text{adv}},\quad\mathcal{L}\big(\boldsymbol{x}_{\text{adv}}+\boldsymbol{\delta}_{d}^{*}\big)\leq\mathcal{L}\big(\boldsymbol{x}_{\text{adv}}\big).
    \label{euqation:loss_decrease}
\end{equation}
Equation~\ref{euqation:loss_decrease} implies that TTR can be viewed as a mapping acting on the LIC system $f$ that never increases the loss. As illustrated in Figure~\ref{fig:shift_contraction_adversarial_region}(b), each sample either preserves its loss or moves to a lower-loss state. Therefore, the adversarial region is contracted rather than shifted. This indicates that TTR improves the intrinsic robustness of the LIC system, rather than creating an illusion. 

Our FTTR is an efficient approximation of TTR. Benefiting from the revealed AAT property, extensive empirical results show that Equation~\ref{euqation:loss_decrease} holds for most adversarial examples. However, since FTTR typically adopts a relatively large strength $s$, it may introduce excessive perturbations. When the adversarial perturbation is small, the negative effects introduced by FTTR may outweigh its benefits. Moreover, for similar reasons, FTTR may degrade benign performance. Therefore, we introduce a conditional branch to determine whether the purifier generated by FTTR provides a gain. If not, we set $\boldsymbol{\delta}_{d}^{(K)}=0$, as shown in Algorithm~\ref{alg:FTTR}. This ensures that Equation~\ref{euqation:loss_decrease} also holds for FTTR. To clearly present the intrinsic robustness of FTTR, we disable this conditional branch in our experiments and evaluate the iteratively optimized purifier.

In summary, this section theoretically established the white-box robustness mechanisms of TTR and our FTTR. The following extensive experiments will empirically validate them.
% Owing to the AAT property of LIC systems, Equation~\ref{euqation:loss_decrease} also empirically holds for FTTR in most cases, despite its single-step approximation of TTR. While FTTR may achieve slightly weaker robustness than fully iterative TTR, it substantially reduces the optimization overhead. This favorable robustness–efficiency trade-off makes FTTR more suitable for practical deployment.

\section{Experiments and Results}
\label{section: Experiments and Results}
\subsection{Setup}
\noindent\textbf{LIC systems.} We evaluate our FTTR across multiple representative LIC systems. Specifically, for untargeted distortion attacks, bitrate collapse attacks, and downstream degradation attacks, we evaluate on two advanced LIC systems DCAE~\cite{lu2025learned} and CCA~\cite{han2024causal}, and three classical models cheng2020-Attn~\cite{cheng2020learned}, bmshj2018-hyperprior~\cite{balle2018variational}, and mbt2018~\cite{minnen2018joint}. For global semantic manipulation, following~\cite{liang2026control}, we evaluate on DCAE, CCA, and HiFi-VRIC~\cite{cai2022high}.

\noindent\textbf{Datasets.} Experiments are conducted on Kodak dataset~\cite{kodak1993}, a standard benchmark for compression, which contains $24$ RGB images of $512\times768$. For the downstream task of classification, we use the NIPS 2017 Adversarial Competition dataset~\footnote{\url{https://github.com/anlthms/nips-2017.git}}.

\noindent\textbf{Comprehensive Adversary Benchmarks.} Prior study~\cite{wu2025adversarial} evaluates only under the gray-box setting. In contrast, we establish a suite of adaptive adversary benchmarks to evaluate the white-box robustness, including BPDA~\cite{athalye2018obfuscated} and Unroll~\cite{andrychowicz2016learning}. BPDA performs standard forward propagation through FTTR but approximates FTTR as an identity mapping during backpropagation, enabling efficient perturbation generation with approximate gradients. In contrast, Unroll explicitly unfolds the optimization process to obtain exact gradients, but incurs substantial computational overhead due to second-order gradients. More details are provided in the appendix. Besides,  we also evaluate FTTR under the gray-box setting.

\noindent\textbf{Implementations.} We set $\lambda_{1}$= $10^{3}$, $\lambda_{2}$=$50$, $s$=$0.04$. Except for downstream defenses, $\lambda_{3}$ is $0$. We adopt 1-step FTTR with $K$=$1$.

\noindent\textbf{Metrics.} For reconstruction quality, we use pixel-level PSNR (dB), structural-level MS-SSIM, and semantic-level LPIPS and CLIP. For compression rate, we use bpp. Note that adversarial examples may cause excessively large bpp, making exact computation infeasible within a reasonable time. Therefore, we report the estimated bpp. For classification tasks, we use top-1 accuracy (\%).

\subsection{To $\ell_{\infty}$ Untargeted Distortion Attacks}
\label{section: experiment_PGD_MSE_attacks}
This experiment evaluates the robustness of FTTR against $\ell_{\infty}$ untargeted distortion attacks based on PGD, where the attack objective is the MSE loss. We compare five settings: (1) \textit{w/o}: both attack and inference are performed on the undefended LIC system $f$. (2) \textit{Noise}: the attack is performed on $f$, while uniform noise with strength 0.04 is added before inference on $f$. (3) \textit{Gray-Box}: the attack is performed on $f$, while inference is performed on the FTTR-equipped LIC system $f_{\mathrm{FTTR}}$. (4) \textit{BPDA}: adaptive white-box attacks are generated against $f_{\mathrm{FTTR}}$ using BPDA, followed by inference on $f_{\mathrm{FTTR}}$. (5) \textit{Unroll}: adaptive white-box attacks are generated against $f_{\mathrm{FTTR}}$ using Unroll, followed by inference on $f_{\mathrm{FTTR}}$. The same protocol is used hereafter and is omitted for brevity. We generate adversarial examples with iterations $T=400$, budgets $\epsilon\in\{6,8,10,12,14,16\}/255$, and step size $\epsilon/T$. The PSNR results are reported in Figure~\ref{fig:experiment_PGD_MSE_PSNR}, with MS-SSIM, LPIPS, and CLIP in the Figure~\ref{fig:experiment_PGD_MSE_MSSSIM_LPIPS_CLIP}. The visualizations are in the Figure~\ref{fig:visualization_DCAE_PGD_16_mse}. At $\epsilon=16/255$, the \textit{w/o} setting yields an average PSNR of only $9.90$ dB, while the FTTR-equipped models maintain $24.58 (+14.68)$ dB and $25.00(+15.10)$ dB under adaptive white-box BPDA and Unroll, respectively. These results demonstrate strong white- and gray-box robustness of FTTR against $\ell_{\infty}$ MSE attacks.

\subsection{To $\ell_{\infty}$ Compression Rate Attacks}
This experiment evaluates the robustness of FTTR against $\ell_{\infty}$ compression rate attacks based on PGD, where the attack objective is the bpp loss. Experimental settings follow Section~\ref{section: experiment_PGD_MSE_attacks}. Results in Figure~\ref{fig:experiment_PGD_bpp_bpp} demonstrate that FTTR achieves large robustness gains against $\ell_{\infty}$ compression rate attacks. On average, at $\epsilon=16/255$, FTTR reduces the bpp from $17.97$ in the w/o setting to $11.83(-6.14)$ under BPDA and $11.57(-6.40)$ under Unroll. The robustness gains vary across LIC systems. Remarkably, for DCAE, the bpp is reduced from 14.29 to 1.102 under BPDA and 1.27 under Unroll. This indicates that certain LIC models can achieve more substantial robustness improvements when combined with FTTR, highlighting the potential of FTTR for trustworthy standardized LIC systems.

\begin{table}[t]
    \centering
    \caption{Reconstruction performance on adversarial examples for high-resolution global semantic manipulation. \textit{FTTR} denotes the \textit{Gray-Box}. The best values are shown in bold.}
    \begin{tabular}{c|c|ccccc}
    \toprule
         \multicolumn{2}{c|}{}&PSNR $\uparrow$&MS-SSIM $\uparrow$&LPIPS $\downarrow$ &CLIP $\uparrow$&bpp $\downarrow$\\
         \midrule

         \multirow{3}{*}[-0ex]{\rotatebox{90}{\textit{DCAE}}}
         &w/o&12.970&0.403&0.687&0.724&1.312 \\
         &Noise&19.146&0.593&0.658&0.830&0.291 \\
         &\cellcolor{gray!13}FTTR&\cellcolor{gray!13}\textbf{21.698}&\cellcolor{gray!13}\textbf{0.717}&\cellcolor{gray!13}\textbf{0.615}&\cellcolor{gray!13}\textbf{0.871}&\cellcolor{gray!13}\textbf{0.141} \\
         \midrule
         
         \multirow{3}{*}[-0ex]{\rotatebox{90}{\textit{CCA}}}
         &w/o&15.704&0.579&0.380&0.786&4.084 \\
         &Noise&21.536&0.834&0.180&0.940&3.746 \\
         &\cellcolor{gray!13}FTTR&\cellcolor{gray!13}\textbf{24.415}&\cellcolor{gray!13}\textbf{0.912}&\cellcolor{gray!13}\textbf{0.076}&\cellcolor{gray!13}\textbf{0.980}&\cellcolor{gray!13}\textbf{1.970} \\
         \midrule
         
         \multirow{3}{*}[-0ex]{\rotatebox{90}{\textit{HiFi-VRIC}}}
         &w/o&12.635&0.390&0.687&0.626&11.148 \\
         &Noise&16.464&0.618&0.530&0.748&9.718 \\
         &\cellcolor{gray!13}FTTR&\cellcolor{gray!13}\textbf{24.002}&\cellcolor{gray!13}\textbf{0.901}&\cellcolor{gray!13}\textbf{0.092}&\cellcolor{gray!13}\textbf{0.982}&\cellcolor{gray!13}\textbf{6.240} \\
    \bottomrule
    \end{tabular}
    
    \label{table:experiments_GSM}
\end{table}

\subsection{To $\ell_{2}$ Joint Rate-Distortion Attacks}
We further evaluate FTTR against $\ell_{2}$ joint R-D attacks based on C\&W attack, where the objective is:
\begin{equation}
    \max -\|\boldsymbol{\delta}_{a}\|_{2}+c\cdot\large[c_{1}\cdot r\large(E\large(\boldsymbol{x}_{\mathrm{adv}}\large)\large) + c_{2}\cdot\mathrm{MSE}\large(f\large(\boldsymbol{x}_{\mathrm{adv}}\large),\boldsymbol{x}\large)\large].
\end{equation}
$c_1$ and $c_2$ are set to $1$ and $500$, respectively. Unlike the standard C\&W attack, we fix three representative $c$ of $500, 750, 1000$. The attack uses Adam with a learning rate of $10^{-3}$ for $1000$ iterations. Results in Figure~\ref{fig:CW_PSNR_BPP} and visualizations in Figure~\ref{fig:DCAE_CW_visualizations} demonstrate the robustness against $\ell_{2}$ attacks. At $c=1000$, on average, PSNR improves from $7.59$ dB in the w/o setting to $26.71(+19.12)$ dB under BPDA, while bpp decreases from $15.19$ to $7.58(-7.61)$.

\begin{figure}
    \centering
    \includegraphics[width=0.94\linewidth]{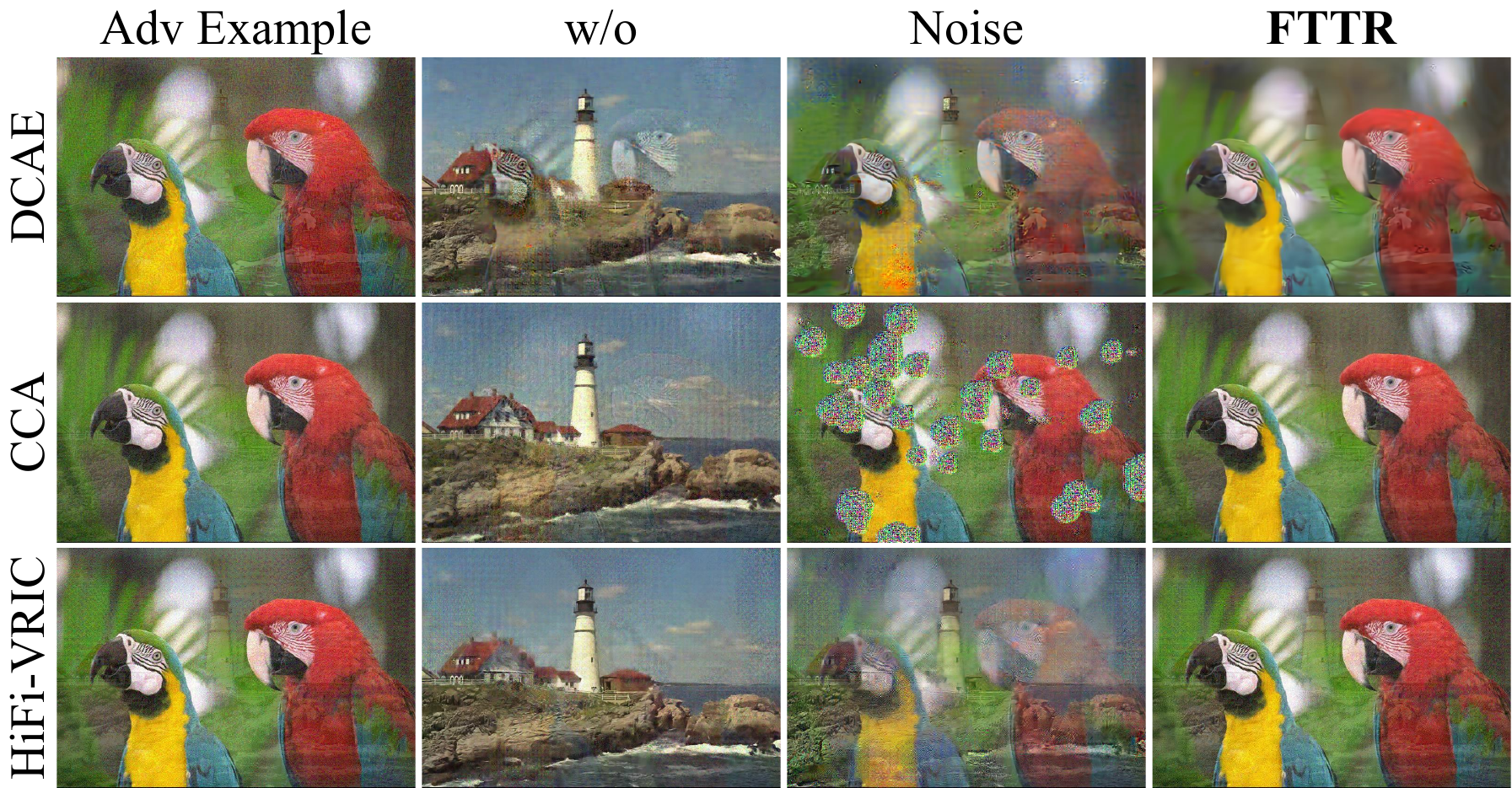}
    \caption{Adversarial examples and reconstructions under global semantic manipulation. \textit{FTTR} denotes the \textit{Gray-Box}.}
    \label{fig:GSM_illustrations}
\end{figure}

\subsection{To Global Semantic Manipulations}
Beyond conventional attacks, we evaluate FTTR against the emerging global semantic manipulation (GSM). We use PGD$^{2}$-GSM~\cite{liang2026control} with $\epsilon=0.10$, while keeping other settings the same as the original paper. Since GSM typically requires tens of thousands of iterations, we only consider the gray-box threat setting here. Quantitative results in Table~\ref{table:experiments_GSM} and visualizations in Figure~\ref{fig:GSM_illustrations} demonstrate that FTTR effectively resists this challenging attack.

\subsection{To LIC-Triggered Downstream Attacks}
We further evaluate FTTR against LIC-triggered downstream classification attacks, which generate adversarial examples that remain benign before LIC processing but cause downstream failures after compression and decompression. The attack is PGD with budget $8/255$, step size $1/255$ and iterations $40$. The classification models include ResNet-50~\cite{he2016deep} and ViT-B/16~\cite{dosovitskiy2020image}. The FTTR strength $s$ is varied from $0.01$ to $0.06$ with an interval of $0.01$. The classification loss is the cross-entropy loss, with $\lambda_{3}$ set to $1$. Results in Figure~\ref{fig:classification} demonstrate that FTTR substantially improves robustness against LIC-triggered downstream attacks. At $s=0.06$, our FTTR boosts the average downstream classification accuracy from $0.4\%$ under \textit{w/o} setting to $28.1\%(+27.7\%)$ under \textit{BPDA} setting.

% \subsection{FTTR on Benign Samples}
% We further evaluate FTTR on benign samples with $K=1$ to $5$. As shown in Figure~\ref{fig:benign}, FTTR causes mild fluctuations at $K=1$ and negligible impact for $K\geq2$, highlighting its remarkable robustness–accuracy trade-off.

\section{Conclusions and Limitations}
Existing robustness studies on LIC focus on gray-box setting, while the white-box robustness of LIC systems remains unexplored. This work presents the first systematic study on the potential of TTR as a white-box defense. We reveal the Asymmetric Adversarial Trajectory (AAT) property of LIC systems that TTR does not require numerous iterations when used for defense purposes. For most LIC systems, a few steps are sufficient. We explain this phenomenon through the hypothesized Tube Model and propose the Fast Test-Time Refinement (FTTR) framework for practical defense deployment. Our analysis shows that TTR and FTTR exploit the Input-as-Label property of LIC systems to contract the adversarial region rather than shift it, leading to genuine robustness improvements. Even with only 1-step FTTR, our experiments demonstrate remarkable white-box robustness gains under various threats. Further optimization of FTTR, such as adaptive strength control, remains an interesting and promising direction for future research.

% \begin{acks}
% \end{acks}

% \section*{Ethics and Privacy Statement}

% This section of your ACM work should discuss the potential societal
% risks that might result from its publication; two to three sentences
% related to the findings of your study, or new advancements made
% possible by their developed methods. The privacy and ethics statement
% should clearly address the broader impacts of their work as it relates
% to the authors' interpretation of privacy, fairness, safety, human
% rights, data sovereignty, or future misuse and any benefit/risk
% trade-off resulting from this research. We acknowledge that some
% papers may have minimal societal risks beyond those considered by
% institutional review boards, and the dimensions considered by any
% review of the user study design or dataset licenses could be provided
% in this statement.

\bibliographystyle{ACM-Reference-Format}
\bibliography{sample-base}

\appendix
\clearpage

\section{Proofs}
\subsection{Deduction of Equation~\ref{equation:reduction_sign_gradient_descent}}
We follow the notations of the original Adam~\cite{kingma2015adam} paper and do not repeat the definitions here. In particular, the variable being optimized is the purifier $\boldsymbol{\delta}_{d}$ rather than the model parameters $\theta$. The parameters are initialized as in the original Adam paper:
\begin{equation}
    \boldsymbol{m}_{0}=\boldsymbol{0},~\boldsymbol{v}_{0}=\boldsymbol{0}.
\end{equation}
The gradient at time step $k$ is:
\begin{equation}
    \boldsymbol{g}_{k}=\nabla_{\boldsymbol{\delta}_{d}^{(k-1)}}\mathcal{L}_{\text{pur}}.
\end{equation}
At time step $k$, Adam updates the purifier $\boldsymbol{\delta}_{d}^{(k)}$ as follows:
\begin{equation}
    \boldsymbol{m}_{k}=\beta_{1}\cdot\boldsymbol{m}_{k-1}+(1-\beta_{1})\cdot\boldsymbol{g}_{k},~\hat{\boldsymbol{m}}_{k}=\frac{\boldsymbol{m}_{k}}{1-\beta_{1}^{k}},
\end{equation}
\begin{equation}
    \boldsymbol{v}_{k}=\beta_{2}\cdot\boldsymbol{v}_{k-1}+(1-\beta_{2})\cdot\boldsymbol{g}_{k}^{2},~\hat{\boldsymbol{v}}_{k}=\frac{\boldsymbol{v}_{k}}{1-\beta_{2}^{k}},
\end{equation}
\begin{equation}
    \boldsymbol{\delta}_{d}^{(k)}=\boldsymbol{\delta}_{d}^{(k-1)}-\alpha\cdot\frac{\hat{\boldsymbol{m}}_{k}}{\sqrt{\hat{\boldsymbol{v}}_{k}}+\epsilon}.
\end{equation}
By setting $k=1$, we obtain:
\begin{equation}
    \boldsymbol{m}_{1}=(1-\beta_{1})\cdot\boldsymbol{g}_{1},~\hat{\boldsymbol{m}}_{1}=\boldsymbol{g}_{1},
\end{equation}
\begin{equation}
    \boldsymbol{v}_{1}=(1-\beta_{2})\cdot\boldsymbol{g}_{1}^2,~\hat{\boldsymbol{v}}_{1}=\boldsymbol{g}_{1}^{2},
\end{equation}
\begin{equation}
    \boldsymbol{\delta}_{d}^{(1)} = \boldsymbol{\delta}_{d}^{(0)}-\alpha\cdot\frac{\boldsymbol{g}_{1}}{|\boldsymbol{g}_{1}|+\epsilon}.
\end{equation}
Given $\boldsymbol{\delta}_{d}^{(0)} = \boldsymbol{0}$ and $\epsilon\ll |\nabla_{\boldsymbol{\delta}_{d}}\mathcal{L}_{\text{pur}}|$, it follows that:
\begin{equation}
    \boldsymbol{\delta}_{d}^{(1)} = -\alpha\cdot\text{sgn}(\boldsymbol{g}_{1}).
\end{equation}

\section{Supplementary Experiments}
\subsection{Details of Adversary Benchmarks}
\noindent\textbf{Unroll.} In white-box scenarios, FTTR can be viewed as a standalone defense module preceding LIC system $f$:
\begin{equation}
\begin{aligned}
    f_{d}=f\circ\mathrm{FTTR}.
\end{aligned}
\end{equation}
For a sample $\boldsymbol{x}$ and $t$-moment perturbation $\boldsymbol{\delta}_{a}^{(t)}$, the attacker performs gradient ascent to optimize it:
\begin{equation}
\begin{aligned}
    \boldsymbol{\delta}_{a}^{(t+1)}=\boldsymbol{\delta}_{a}^{(t)}+\alpha\cdot\nabla_{\boldsymbol{\delta}_{a}^{(t)}}\mathcal{L}(f_{d}(\boldsymbol{x}+\boldsymbol{\delta}_{a}^{(t)}),\boldsymbol{x}).
\label{equation:attack_update}
\end{aligned}
\end{equation}
We expand the gradient in Equation~\ref{equation:attack_update} as follows:
\begin{equation}
\begin{aligned}
    &\nabla_{\boldsymbol{\delta}_{a}^{(t)}}\mathcal{L}\big(f_{d}\big(\boldsymbol{x}+\boldsymbol{\delta}_{a}^{(t)}\big),\boldsymbol{x}\big)
    \\=&\nabla_{\boldsymbol{\delta}_{a}^{(t)}}\mathcal{L}\big(f\big(\mathrm{FTTR}\big(\boldsymbol{x}+\boldsymbol{\delta}_{a}^{(t)}\big)\big),\boldsymbol{x}\big)
    \\=&\nabla_{\boldsymbol{\delta}_{a}^{(t)}}\mathcal{L}\big(f\big(\boldsymbol{x}+\boldsymbol{\delta}_{a}^{(t)}+\boldsymbol{\delta}_{d}^{(K)}\big),\boldsymbol{x}\big)
    \\=&\frac{\partial\mathcal{L}}{\partial f}\cdot\frac{\partial f}{ \partial\big(\boldsymbol{x}+\boldsymbol{\delta}_{a}^{(t)}+\boldsymbol{\delta}_{d}^{(K)}\big)}\cdot\frac{\partial\big(\boldsymbol{x}+\boldsymbol{\delta}_{a}^{(t)}+\boldsymbol{\delta}_{d}^{(K)}\big)}{\partial\boldsymbol{\delta}_{a}^{(t)}}
    \\=&\frac{\partial\mathcal{L}}{\partial f}\cdot\frac{\partial f}{ \partial\big(\boldsymbol{x}+\boldsymbol{\delta}_{a}^{(t)}+\boldsymbol{\delta}_{d}^{(K)}\big)}\cdot\big(\boldsymbol{I}+\frac{\partial\boldsymbol{\delta}^{(K)}_{d}}{\partial\boldsymbol{\delta}^{(t)}_{a}}\big).
\label{equation:expand}
\end{aligned}
\end{equation}
According to Algorithm~\ref{alg:FTTR}, $\boldsymbol{\delta}^{(K)}_{d}$ is obtained through an iterative optimization process with respect to $\boldsymbol{\delta}^{(t)}_{a}$. In our experiments, $K=1$, and thus we have:
\begin{equation}
\begin{aligned}
    \boldsymbol{\delta}_{d}^{(1)}=\text{Clip}_{(-s,s)}\big(\boldsymbol{\delta}_{d}^{(0)}-s\cdot\text{sgn}\big(\nabla_{\boldsymbol{\delta}_{d}^{(0)}}\mathcal{L}_{\text{FTTR}}^{(1)}\big)\big).
\end{aligned}
\end{equation}
In our experiments, we adopt the straight-through estimation (STE) for $\mathrm{Clip}(\cdot)$ and $\mathrm{sgn}(\cdot)$ by assuming their gradients are identity mappings. Therefore, the partial derivative of $\boldsymbol{\delta}_{d}^{(1)}$ with respect to $\boldsymbol{\delta}_{a}^{(t)}$ is:
\begin{equation}
\begin{aligned}
    \frac{\partial\boldsymbol{\delta}_{d}^{(1)}}{\partial\boldsymbol{\delta}_{a}^{(t)}}=-s\cdot\frac{\partial}{\partial\boldsymbol{\delta}_{a}^{(t)}}\big(\nabla_{\boldsymbol{\delta}_{d}^{(0)}}\mathcal{L}_{\text{FTTR}}^{(1)}\big).
\label{equation:second_order_derivatives}
\end{aligned}
\end{equation}
Equation~\ref{equation:second_order_derivatives}, i.e., the second term inside the last summation parentheses in Equation~\ref{equation:expand}, requires second-order derivatives. This is computationally expensive in practice. We implement the second-order derivatives using the PyTorch framework, where they are internally computed via Hessian-vector products (HVPs). This procedure is commonly referred to as unrolled optimization in previous literature~\cite{andrychowicz2016learning}. It represents a strong adaptive white-box attack against FTTR, as it obtains the complete gradient information through the defense module. Although this process is computationally complex, we still evaluate FTTR against this strong adversary as a benchmark in this work.

\noindent\textbf{BPDA.} As unrolling incurs substantial computational overhead, we further explore BPDA~\cite{athalye2018obfuscated} as a more efficient alternative. Specifically, BPDA approximates the backward pass of FTTR with an identity mapping while preserving its original forward computation. Consequently, the attack gradient is approximated as:
\begin{equation}
\begin{aligned}
    &\nabla_{\boldsymbol{\delta}_{a}^{(t)}}\mathcal{L}\big(f\big(\mathrm{FTTR}\big(\boldsymbol{x}+\boldsymbol{\delta}_{a}^{(t)}\big)\big),\boldsymbol{x}\big)
    \\=&\frac{\partial\mathcal{L}}{\partial f}\cdot\frac{\partial f}{\partial\mathrm{FTTR}}\cdot\frac{\partial\mathrm{FTTR}}{\partial\big(\boldsymbol{x}+\boldsymbol{\delta}_{a}^{(t)}\big)}
    \\\approx&\frac{\partial\mathcal{L}}{\partial f}\cdot\frac{\partial f}{\partial\mathrm{FTTR}}\cdot\boldsymbol{I},
\end{aligned}
\end{equation}
thereby bypassing the expensive second-order derivatives required by unrolled optimization. Although BPDA provides only an approximate gradient, it has been widely adopted for evaluating defenses involving non-differentiable or iterative modules. In this work, we evaluate FTTR against both Unroll and BPDA to provide a comprehensive assessment under adaptive white-box attacks.

\noindent\textbf{Gray-Box Adversary Benchmarks.} For gray-box adversary benchmarks, the attacker crafts adversarial examples based solely on the undefended LIC system $f$, i.e.,
\begin{equation}
\begin{aligned}
    \boldsymbol{\delta}_{a}=A\big(f, \boldsymbol{x}\big).
\end{aligned}
\end{equation}

\begin{figure*}[t]
    \centering
    \includegraphics[width=1.0\linewidth]{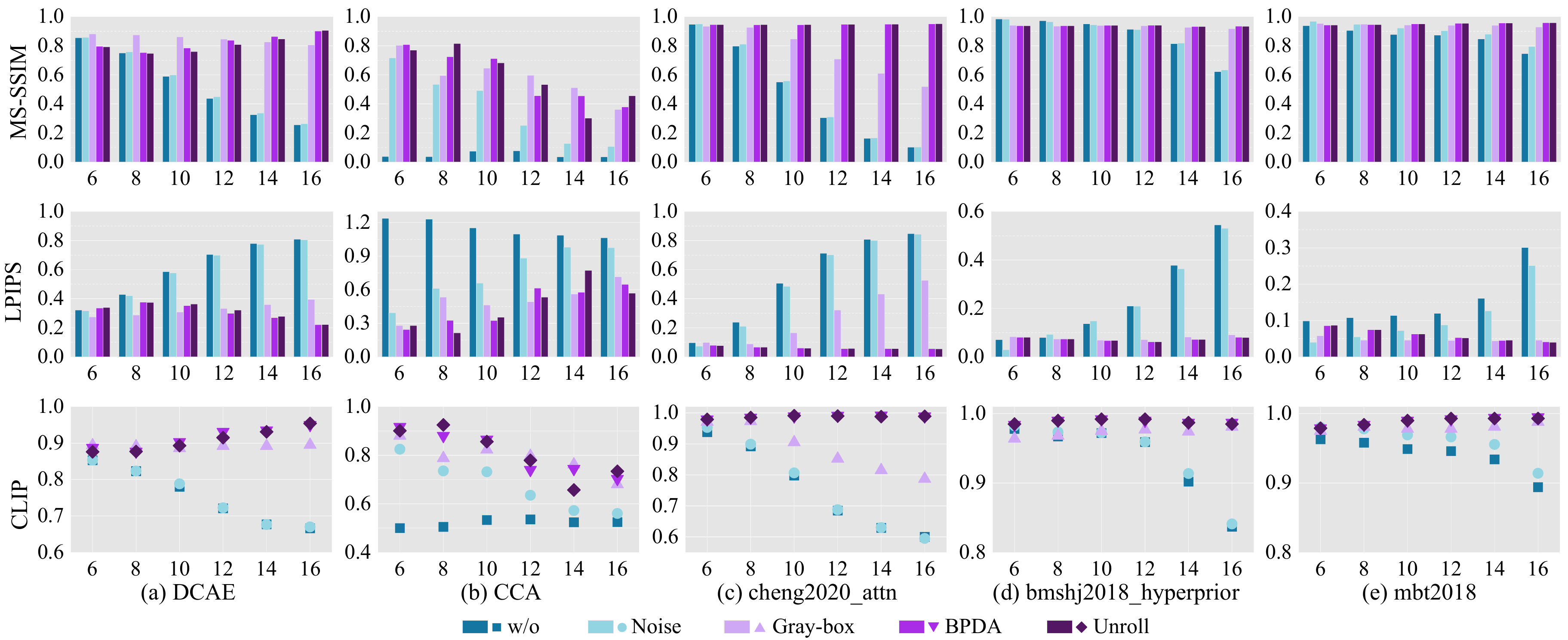}
    \caption{MS-SSIM $\uparrow$, LPIPS $\downarrow$, and CLIP $\uparrow$ of PGD-attacked reconstructions under MSE. X-axis indicates perturbation budgets.}
    \label{fig:experiment_PGD_MSE_MSSSIM_LPIPS_CLIP}
\end{figure*}

\begin{figure*}[t]
    \centering
    \includegraphics[width=0.83\linewidth]{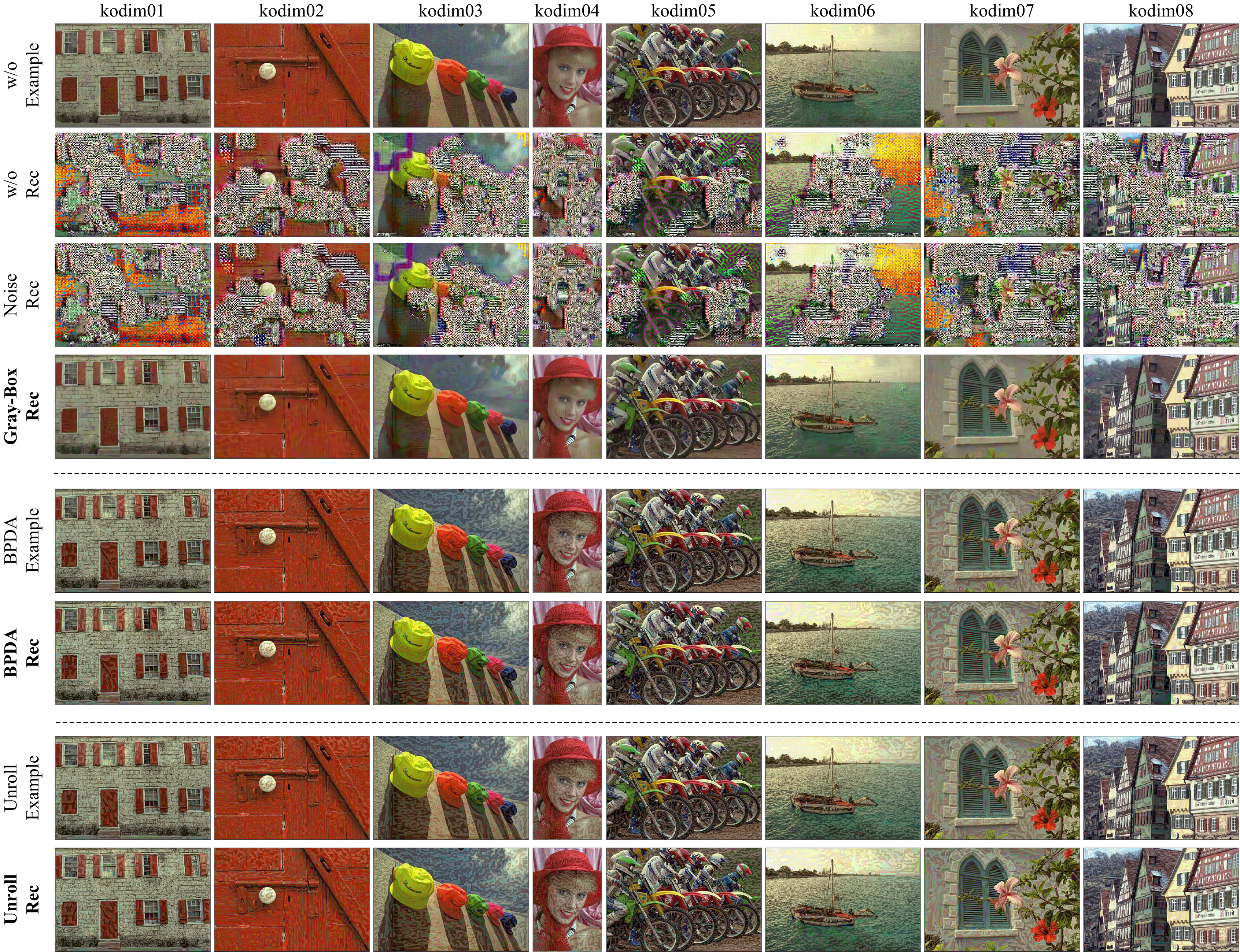}
    \caption{Visualizations of PGD-attacked adversarial examples and their reconstructions under MSE on DCAE. \textit{Example} indicates adversarial examples. \textit{Rec} indicates reconstructions.}
    \label{fig:visualization_DCAE_PGD_16_mse}
\end{figure*}

\begin{figure*}[t]
    \centering
    \includegraphics[width=0.68\linewidth]{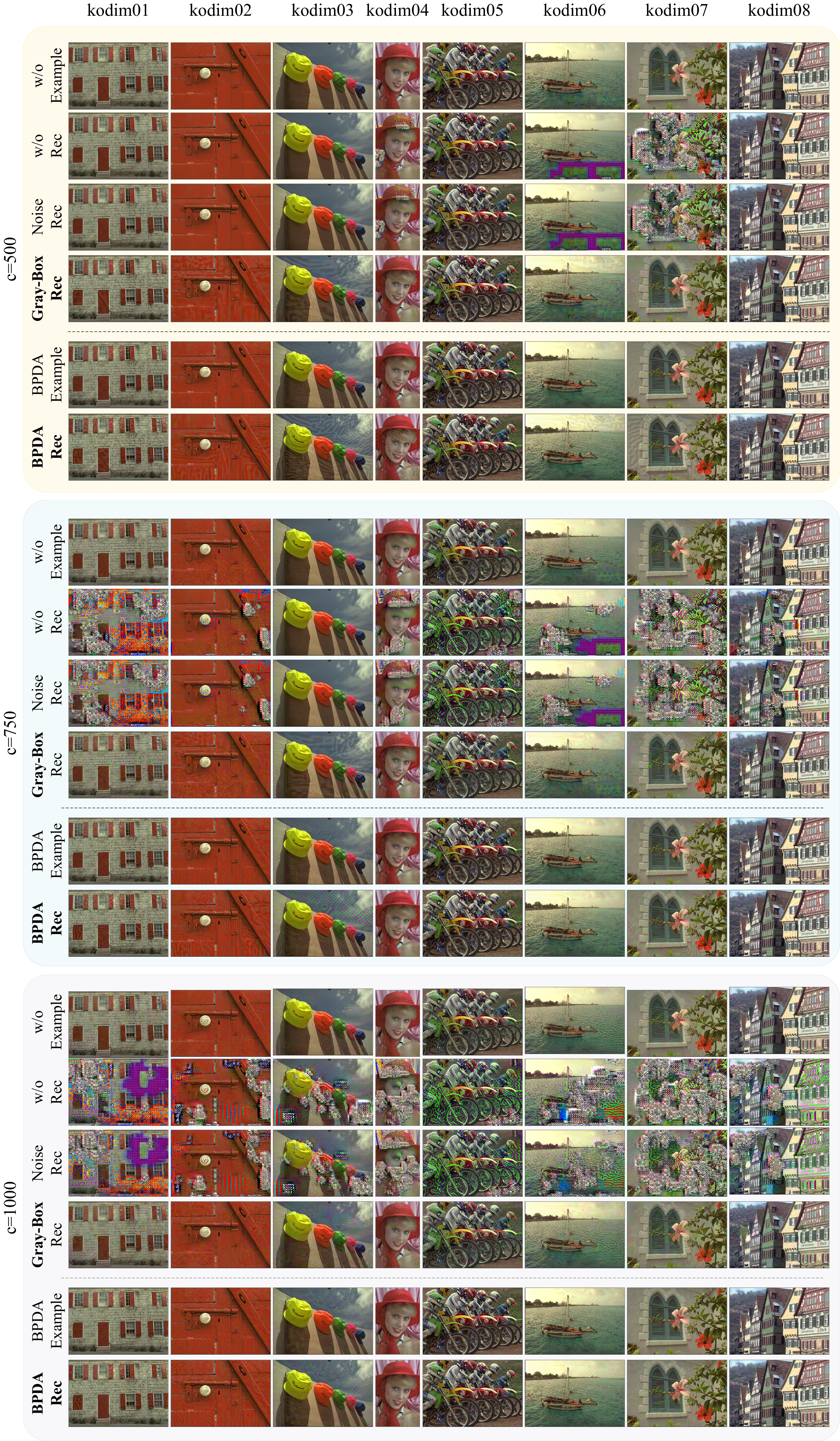}
    \caption{Visualizations of $\ell_{2}$ C\&W-attacked adversarial examples and their reconstructions under joint objective of MSE and bpp on DCAE. \textit{Example} indicates adversarial examples. \textit{Rec} indicates reconstructions.}
    \label{fig:DCAE_CW_visualizations}
\end{figure*}

\end{document}